\documentclass[sn-mathphys,Numbered]{sn-jnl}

\usepackage{graphicx}%
\usepackage{multirow}%
\usepackage{amsmath,amssymb,amsfonts}%
\usepackage{amsthm}%
\usepackage{mathrsfs}%
\usepackage[title]{appendix}%
\usepackage{xcolor}%
\usepackage{textcomp}%
\usepackage{manyfoot}%
\usepackage{booktabs}%
\usepackage{algorithm}%
\usepackage{algorithmicx}%
\usepackage{algpseudocode}%
\usepackage{listings}%
\usepackage{xspace}
\usepackage{numprint}
\usepackage{makecell}
\usepackage{dirtytalk}
\usepackage[normalem]{ulem}
\usepackage{booktabs}

\theoremstyle{thmstyleone}%
\theoremstyle{thmstyletwo}%

\theoremstyle{thmstylethree}%

\newcommand{\SUP}{\texttt{SUPPORTS}\xspace}

\newcommand{\REF}{\texttt{REFUTES}\xspace}

\newcommand{\NEI}{\texttt{NEI}\xspace}
\newcommand{\FAIL}{\texttt{FAILED}\xspace}

\newcommand{\Anserini}{\textsc{Anserini}\xspace}
\newcommand{\BART}{\textsc{BART}\xspace}
\newcommand{\BERT}{\textsc{Bert}\xspace}
\newcommand{\ColBERT}{\textsc{ColBERT}\xspace}
\newcommand{\ColBERTvT}{\textsc{ColBERTv2}\xspace}
\newcommand{\Czert}{\textsc{Czert}\xspace}
\newcommand{\DrQA}{\textsc{DrQA}\xspace}
\newcommand{\HerBERT}{\textsc{HerBERT}\xspace}
\newcommand{\MBART}{\textsc{MBART}\xspace}
\newcommand{\MBERT}{\textsc{M-BERT}\xspace}
\newcommand{\MTFIVE}{\textsc{mT5}\xspace}
\newcommand{\Polbert}{\textsc{Polbert}\xspace}
\newcommand{\PoLitBert}{\textsc{PoLitBert}\xspace}
\newcommand{\RobeCzech}{\textsc{RobeCzech}\xspace}
\newcommand{\RoBERTa}{\textsc{RoBERTa}\xspace}

\newcommand{\SlavicBERT}{\textsc{SlavicBERT}\xspace}
\newcommand{\SlovakBERT}{\textsc{SlovakBERT}\xspace}

\newcommand{\UMTFIVE}{\textsc{UmT5}\xspace}
\newcommand{\XLMR}{\textsc{XLM-RoBERTa}\xspace}

\newcommand{\FCZ}{\textsc{CsFEVER}\xspace}
\newcommand{\FDAN}{\textsc{DanFEVER}\xspace}
\newcommand{\FEN}{\textsc{EnFEVER}\xspace}
\newcommand{\FEVER}{\textsc{FEVER}\xspace}
\newcommand{\FEVERNLI}{\textsc{FEVER-NLI}\xspace}
\newcommand{\FCZNLI}{\textsc{CsFEVER-NLI}\xspace}
\newcommand{\FENNLI}{\textsc{EnFEVER-NLI}\xspace}

\newcommand{\QATD}{\textsc{QA2D}\xspace}

\newcommand{\QACG}{\textsc{QACG}\xspace}
\newcommand{\QACGCS}{\textsc{QACG-cs}\xspace}
\newcommand{\QACGEN}{\textsc{QACG-en}\xspace}
\newcommand{\QACGPL}{\textsc{QACG-pl}\xspace}
\newcommand{\QACGSK}{\textsc{QACG-sk}\xspace}
\newcommand{\QACGMIX}{\textsc{QACG-mix}\xspace}
\newcommand{\QACGSUM}{\textsc{QACG-sum}\xspace}
\newcommand{\QACGCSFS}{\textsc{QACG-cs-f}\xspace}
\newcommand{\QACGENFS}{\textsc{QACG-en-f}\xspace}

\newcommand{\CS}{\texttt{cs}\xspace}
\newcommand{\EN}{\texttt{en}\xspace}
\newcommand{\PL}{\texttt{pl}\xspace}
\newcommand{\SK}{\texttt{sk}\xspace}
\newcommand{\MIX}{\texttt{mix}\xspace}

\newcommand{\SUM}{\texttt{sum}\xspace}

\newcommand{\ColBERTANS}{\textsc{ColBERT\textsubscript{ANS}}\xspace}
\newcommand{\ColBERTNLI}{\textsc{ColBERT\textsubscript{NLI}}\xspace}

\newcommand{\train}{\texttt{train}\xspace}
\newcommand{\dev}{\texttt{dev}\xspace}
\newcommand{\test}{\texttt{test}\xspace}
\newcommand{\total}{\texttt{total}\xspace}

\newcommand{\red}[1]{{\color{red}#1}}
\newcommand{\blue}[1]{{\color{blue}#1}}
\newcommand{\orange}[1]{{\color{orange}#1}}

\begin{document}

\title[Article Title]{Pipeline and Dataset Generation for Automated Fact-checking in Almost Any Language}


\author*[1]{\fnm{Jan} \sur{Drchal}}\email{drchajan@fel.cvut.cz}

\author[1]{\fnm{Herbert} \sur{Ullrich}}\email{ullriher@fel.cvut.cz}

\author[1]{\fnm{Tom\'{a}\v{s}} \sur{Mlyn\'{a}\v{r}}}\email{mlynatom@fel.cvut.cz}


\author[2]{\fnm{V\'{a}clav} \sur{Moravec}}\email{vaclav.moravec@fsv.cuni.cz}

\affil*[1]{\orgdiv{Artificial Intelligence Center}, \orgname{Faculty of Electrical Engineering, Czech Technical University in Prague}, \orgaddress{\street{Charles Square 13}, \city{Prague~2}, \postcode{120 00}, \country{Czech Republic}}}

\affil[2]{\orgdiv{Department of Journalism}, \orgname{Faculty of Social Sciences, Charles University}, \orgaddress{\street{Smetanovo n\'{a}b\v{r}e\v{z}\'{i} 6}, \city{Prague~1}, \postcode{110 01},  \country{Czech Republic}}}



\abstract{
This article presents a pipeline for automated fact-checking leveraging publicly available Language Models and data.
The objective is to assess the accuracy of textual claims using evidence from a ground-truth \textit{evidence corpus}.
The pipeline consists of two main modules -- the \textit{evidence retrieval} and the \textit{claim veracity evaluation}.
Our primary focus is on the ease of deployment in various languages that remain unexplored in the field of automated fact-checking.
Unlike most similar pipelines, which work with evidence sentences, our pipeline processes data on a paragraph level, simplifying the overall architecture and data requirements. 
Given the high cost of annotating language-specific fact-checking training data, our solution builds on the Question Answering for Claim Generation (\QACG) method, which we adapt and use to generate the data for all models of the pipeline.
Our strategy enables the introduction of new languages through machine translation of only two fixed datasets of moderate size.
Subsequently, any number of training samples can be generated based on an evidence corpus in the target language.
We provide open access to all data and fine-tuned models for Czech, English, Polish, and Slovak pipelines, as well as to our codebase that may be used to reproduce the results.\footnote{\url{https://github.com/aic-factcheck/multilingual-fact-checking}}~~We comprehensively evaluate the pipelines for all four languages, including human annotations and per-sample difficulty assessment using Pointwise $\mathcal{V}$-information.
The presented experiments are based on full Wikipedia snapshots to promote reproducibility.
To facilitate implementation and user interaction, we develop the FactSearch application featuring the proposed pipeline and the preliminary feedback on its performance.
}

\keywords{automated fact-checking, evidence retrieval, claim veracity evaluation, QACG}



\maketitle
\section{Introduction}\label{sec:intro}

Recent NLP research has extensively covered the case of \textit{automated fact-checking}~\cite{guo-etal-2022-survey}.
The task is to evaluate \textit{veracity} of a \textit{factual claim} as well as to provide \textit{evidence} supporting or refuting the claim.
The state-of-the-art \textit{automated fact-checking} deals mainly with two paradigms: the \textit{open-world fact-checking}~\cite{chen2023complex} where the evidence is collected using publicly available Internet search engines, and the \textit{closed-world fact-checking} in which the claim is inspected with respect to a \textit{evidence corpus}.
While the first approach has the advantage of practically unlimited information sources, its weakness generally lies in the low reliability of the public data. 
This paper deals with the \textit{closed-world} paradigm, although we do not specifically focus on the data reliability but rather on the construction of the fact-checking pipeline.
We aim to provide reproducible results and opt for the publicly available Wikipedia data. 

The challenge of verifying factual claims with references to an \textit{evidence corpus} has been modeled as a multiple-step task where the key modules constitute \textit{evidence retrieval} and \textit{veracity labeling}~\cite{thorne2018fever}.
A number of approaches were benchmarked~\cite{thorne2018shared} and addressed with solutions steadily reaching maturity~\cite{guo-etal-2022-survey,chen2023complex}.

\textit{Evidence retrieval} is closely connected to the broader concept of \textit{document retrieval}.
While document retrieval typically focuses on retrieving semantically similar documents or those containing query keywords, evidence retrieval goes beyond this scope.
Its primary goal is to gather evidence for assessing the veracity of a claim.
The \textit{veracity labeling} solves the Natural Language Inference (NLI) task by classifying the claim and the retrieved evidence into classes such as \SUP, \REF, and \NEI (Not Enough Information).~\footnote{While different classification schemes exist, we work with these three classes as defined in \FEVER~\cite{thorne2018fever} throughout the paper.}

With well-performing English solutions on the rise, solutions in other languages and the multilingual setting fall behind~\cite{norregaard-derczynski-2021-danfever,ullrich2023csfever}.
The main bottleneck in the multilingual reproduction of studies that have scored well in English is often the absence of large-scale datasets in other languages~\cite{ullrich2023csfever}.
While the use of Machine Translation methods to directly obtain the fact-checking data has been revealed to produce large proportions of noise~\cite{ullrich2023csfever}, there is another solution to obtain the fact-checking samples in arbitrary languages: generating claims synthetically from their evidence.
The Question Answering for Claim Generation (\QACG) has shown promising results in English~\cite{pan2021zero}, which encourages its reproduction in lower-resource languages where the need for synthesizing data is dire.

Furthermore, the commonly used FEVER~\cite{thorne2018fever} benchmark and its numerous derivatives rely on the corpus of \textit{lead page sections} from Wikipedia snapshot to retrieve its \textit{sentences} for evidence.
This approach does not employ the full extent of data available, and the efficiency of modern methods encourages the use of larger chunks of text to retrieve, providing a more precise semantic understanding than an out-of-context sentence.  

In this article, we attempt to address these challenges, proposing a scheme of claim-generation-driven dataset synthesis, using the \textit{complete} Wikipedia corpus chunked into \textit{paragraphs}.
The shift from more common \textit{sentence}-level processing~\cite{thorne2018fever} to the \textit{paragraph}-level processing is motivated by the increased ability of state-of-the-art large language models to process longer contexts, but also by the overall simplification of the processing pipeline. 
We have already advocated for this approach in our previous work~\cite{ullrich2023csfever}.

Along with English, we study the scheme's efficiency in the case of West Slavic languages (Czech, Polish, and Slovak) as well as in the multilingual scenario, evaluate the quality of data, and establish a functioning fact-checking pipeline in all four languages on top of it.

The contributions of this paper are as follows:
\begin{itemize}
    \item We build a fact-checking pipeline aimed to be easily adaptable to any language using limited computational resources.
    \item The unavailability of data in the target language is overcome using a synthetically generated fact-checking dataset.
    \item Our approach builds on \QACG data-generation method~\cite{pan2021zero} updated for working on a paragraph-level; we also modify the \REF claim generation, replacing the monolingual \textsc{sense2vec} model~\cite{trask2015sense2vec} by a significantly more accessible \textit{named entity recognition} approach.
    \item We build pipelines for four languages: Czech, English, Polish, and Slovak, which involves datasets and models that we make publicly available:
    \begin{itemize}
        \item SQuAD-cs --- a machine-translated Czech version of the SQuAD~\cite{rajpurkar2016squad} \textit{question answering} dataset.
        \item Machine-translated Czech, Polish, and Slovak versions of the \QATD dataset~\cite{demszky2018transforming}, aimed to train models transforming \textit{questions} into corresponding \textit{declarative} sentences.
        \item \textit{Question generation} (QG) and \textit{claim generation} (CG) models trained using the SQuAD and \QATD data for all four languages.
        \item Complete fact-checking Wikipedia-based \QACG datasets generated for all four languages.
        \item \textit{Evidence retrieval} (ER) and \textit{natural language inference} (NLI) models trained using the generated data for all four languages.
The publicly available ER models are based on the state-of-the-art \ColBERTvT~\cite{santhanam2022colbertv2} \textit{document retrieval} method.
We also calibrate the NLI models using \textit{temperature scaling}~\cite{guo2017calibration} to provide more realistic confidence estimations.
        \item Implementation (source code) of the pipeline for all four languages.
    \end{itemize}
    \item We experiment with different approaches to improve the \textit{evidence retrieval} proposing two methods: \ColBERTANS (evidence filtering) and \ColBERTNLI (evidence reranking).
    \item We evaluate the pipeline not only using the common quantitative metrics but also by means of human annotations.
While authors of \QACG focused solely on evaluating the data on the NLI task, we extend our analysis to the ER as well.
    \item We estimate the \textit{difficulty} of our data w.r.t. the NLI classifier analysing the Pointwise $\mathcal{V}$-information~\cite{ethayarajh2022pvi}.
    \item We provide a prototype web application (\textit{FactSearch}) running the pipelines and user feedback report.
The \textit{FactSearch} source code is publicly available.
\end{itemize}

The paper is structured as follows.
Section~\ref{sec:related} explores the related work and the current state of the art in related tasks.
Section~\ref{sec:approach} describes our proposed fact-checking pipeline.
Section~\ref{sec:data} describes all corpora, data used to train \QACG models, as well as the \QACG models and generated \QACG datasets for Czech, English, Polish, and Slovak.
In Section~\ref{sec:exp}, we train the pipeline models and evaluate the pipeline both quantitively and qualitatively.
Section~\ref{sec:fsearch} deals with implementing our fact-checking pipeline in \textit{FactSearch} application and provides user feedback.
The last Section~\ref{sec:conclusion} concludes and gives a future research overview.
\section{Related Work}\label{sec:related}

Despite the abundance of fact verification methods and data available in English~\cite{guo-etal-2022-survey}, only several recent publications have covered the topic of multilingual fact-checking.
X-Fact~\cite{gupta-srikumar-2021-x} presents fact-checking data in 25 languages, using a crawled website text from Google's fact-check explorer as evidence, making the data appropriate for the NLI task, but not for retrieval, as the evidence string is not linked to a retrievable document.

Demagog~\cite{priban-etal-2019-machine} data in West Slavic languages (Czech, Slovak, Polish) feature claim, veracity verdict, its human-annotated rationale, and metadata, such as the political affiliation of the speaker, making it more appropriate for statistical analysis, as the rationale can not be converted to gold ER or NLI data.

An important dataset in English featuring gold data for both veracity labeling and retrieval step is the FEVER~\cite{thorne2018fever} dataset (referred to as \FEN in this article for disambiguation).
\FEN operates with an enclosed evidence corpus of Wikipedia abstracts (first section of each article).
It publishes 185K human-annotated claims along with their veracity verdict w.r.t. Wikipedia and pointers to corpus sentences that were used to infer each label.
In~\cite{norregaard-derczynski-2021-danfever}, a smaller Danish variant of \FEVER, referred to as \FDAN, is detailed.

Our previous publication~\cite{ullrich2023csfever} proposes a scheme of synthesizing fact-checking data in a multilingual setting by machine translation of the English dataset and proceeds to obtain \FCZ{} -- a full-pipeline fact-checking dataset using the fact-checking information featured in \FEN{} and transferring it to a document-aligned Czech Wikipedia corpus.
Omitting data points requiring a document missing in the Czech Wikipedia, the \FCZ{} delivers 127K labeled claims, translated from \FEN.
The article also describes the caveats of this scheme, finds a high level of noise in the data, and motivates the exploration of methods to generate claims directly based on retrievable evidence using a synthetic claim generation scheme.

Question Answering for Claim Generation (\QACG)~\cite{pan2021zero} generates such claims by using a \textit{question generation} and \textit{claim generation} pipeline on a Wikipedia-based evidence corpus -- \texttt{SUPPORT}ed claims are formed as a declarative sentence directly from the question and its answer, \texttt{REFUTE}d ones perturb the claims by altering evidence facts with retrieved randomized similar facts retrieved using \textsc{sense2vec}~\cite{trask2015sense2vec}.
\NEI claims are emulated by augmenting the claim facts with extra contexts on top of the provided evidence.
\QACG was already used in the context of fact-checking scientific articles~\cite{wright2022generating}.
The latter scheme has so far only been used for NLI fine-tuning in English. Our primary focus is evidence retrieval, which motivates the exploration of similar methods to synthesize gold data for the ER task on the fact-checking pipeline and to reliably generate claims from gold evidence in languages other than English.

\subsection{Information Retrieval}

Traditionally, fact-checking pipelines consist of two-level retrieval: \textit{document retrieval} and \textit{sentence selection}~\cite{thorne2018shared, thorne2019fever2}. 
We advocate for omitting the latter step and forming the domain of all documents on the granularity of corpus \textit{paragraphs} rather than its whole articles~\cite{ullrich2023csfever}.
The main advantage over retrieving sentences alone is the additional context the whole paragraph provides, often needed to interpret the evidence semantics correctly (e.g., resolving coreferences).

While lexical methods such as BM-25 implemented by \Anserini~\cite{yang2017anserini} or TF-IDF-based \DrQA~\cite{chen2017drqa} yield a strong baseline at a meager computational cost, the current state of the art in this task in standard benchmarks is being held by methods based on language models~\cite{beir}.

The language model-based retrieval methods use mostly two approaches: two-tower architectures~\cite{karpukhin2020dense, xiong2021approximate} and cross-encoders~\cite{reimers2019sbert}.
The two-tower model features separate neural network towers for independently processing query and document embeddings, computing their semantic similarity afterward (e.g., using cosine similarity).
In contrast, the cross-encoder model employs a single language model that processes query and document simultaneously, capturing a joint understanding throughout the entire forward pass.
Although the two-tower approach enables precomputing embeddings for all corpus documents before conducting a search, leading to high scalability, it tends to lag in retrieval quality compared to the superior performance of cross-encoders.

Our previous research has shown \ColBERT~\cite{khattab2020colbert} to work well for the retrieval task in less-resourced languages~\cite{ullrich2023csfever}. 
Unlike other language model-based methods centered around full document embeddings, \ColBERT uses the \textit{late-iteraction} architecture working with embeddings at the granularity of individual tokens, taking best from two-tower models and cross-encoders.
The model is trained using pairwise softmax cross-entropy loss computed over \textit{(query, positive document, negative document)} triplets.

\ColBERTvT~\cite{santhanam2022colbertv2} improves upon \ColBERT performance by working with $n$-way tuples composed of the positive document and a sequence of hard negatives ordered by decreasing score given by the bootstrapping retrieval model (e.g., the original \ColBERT).
Another improvement lies in introducing a highly efficient compression mechanism to store the embeddings, significantly reducing the memory footprint.

\subsection{Natural Language Inference (NLI)}
Deciding claim veracity given pre-retrieved textual evidence has been traditionally modeled as an NLI task~\cite{thorne2018shared,nie2019combining,ullrich2023csfever} -- a ternary classification of the inference relationship between two texts (\SUP, \REF, \NEI).
Among the NLI datasets, one frequently used for its fact-checking application is the \FENNLI obtained in~\cite{nie2019combining} as a way to interpret \FEN data as \textit{context, query, label} triples.

\cite{ullrich2023csfever} proceeds to translate this data into Czech using the DeepL service, publishing the resulting \FCZNLI.
It thoroughly benchmarks the available models, arriving at a cross-encoder~\cite{reimers2019sbert} architecture and multilingual transformer models.

\subsection{Multilingual Models}
Currently, multiple tasks in the NLP field are seeing the advent of Large Language Models such as GPT-4~\cite{gpt4}, LLaMA~\cite{touvron2023llama}, and Mistral~\cite{jiang2023mistral}, and the paradigm is shifting from fine-tuning models on task-specific data towards few-shot learning and chain-of-thought prompting~\cite{zhao2023survey}.
While the proprietary models such as GPT-4 are offered only as a black box, obscuring the derived research reproducibility, the open-source LLMs strive to match their English-speaking performance using much fewer model parameters.
This is generally achieved, among other strategies, by filtering out the non-English data from their training corpora. Therefore, a multilingual language model of this magnitude that would justify its computational cost is yet to be trained\footnote{As of December 2023.}.

For that reason, we examine solutions based on smaller models pre-trained on general tasks and multilingual data, such as \MBERT~\cite{devlin2019bert} (trained on Wikipedia in 104 languages using Masked Language Modeling (MLM) and Next Sentence Prediction tasks), \XLMR~\cite{conneau2020unsupervised} (pre-trained only on MLM) has reached the state-of-the-art performance on our previous Czech fact-checking datasets~\cite{ullrich2023csfever}. 
In West Slavic languages we explore in this article, \Czert~\cite{sido2021czert} was trained using the \BERT training procedure, \RobeCzech~\cite{straka2021robeczech} used the \RoBERTa~\cite{liu2019roberta} pre-training tasks on Czech data.
SlovakBERT~\cite{pikuliak2022slovakbert} is a similar monolingual model for Slovak.
\HerBERT~\cite{rybak2020herbert}, \Polbert~\cite{kleczek2020polbert}, and \PoLitBert~\footnote{See \url{https://github.com/Ermlab/PoLitBert}.} are similar models for Polish.
\SlavicBERT~\cite{arkhipov2019tuning} is a multilingual model pretrained for Russian, Bulgarian, Czech, and Polish.

In the following chapters, we proceed to use these findings and the current state of the art on related NLP tasks to build a pipeline and a way of acquiring data easily adaptable to various languages, testing the approach on West Slavic languages.
\section{Our Approach}\label{sec:approach}
This section describes our approach to the automated fact-checking pipeline.
We describe the pipeline architecture and updated \QACG procedure generating training data for the pipeline models.

\subsection{Pipeline Overview}
\label{sec:pipeline_overview}

Our claim fact-checking pipeline aims to fulfill several design goals:
\begin{enumerate}
    \item The pipeline design allows relatively easy implementation of other languages.
All models involved can be trained using datasets of moderate size. If the needed dataset is unavailable for the target language, one can employ machine translation using reasonably limited resources.

    \item Unlike in original \FEVER and according to our previous work~\cite{ullrich2023csfever}, we opted to work on a paragraph level instead of sentence level, which means that each piece of \textit{evidence} (an \textit{evidence document}) corresponds to a single Wikipedia paragraph.  
The motivation of the paragraph-level approach lies mainly in the extended context of the paragraph when compared to an individual sentence.
Also, the longer input size is less of a problem for the modern LLMs.
Overall, the pipeline is notably simplified by omitting the \textit{sentence selection}~\cite{thorne2018fever} step.

    \item Based on user feedback during the development, and given that our focus is not on multi-hop fact-checking, we have decided not to provide overall claim veracity evaluation. Instead, our pipeline reports \textit{per-evidence document veracity} hints.
While giving overall verdict (typically \SUP, \REF, or \NEI) has been a convention since the early pipelines~\cite{thorne2018fever}, we have found that due to the limited accuracy of NLI models, it is better to provide a detailed per-document veracity assessment, shifting towards the \textit{assistive} tool approach, where users get a better idea of the decision-making process, instead of dealing with black-box answers only.
Note that verdict explainability is critical -- an unexplained black-box veracity classification has little use for the actual human fact-checkers~\cite{hrckova2022automated}.
\end{enumerate}

\begin{figure}
    \centering
    \includegraphics[width=\textwidth]{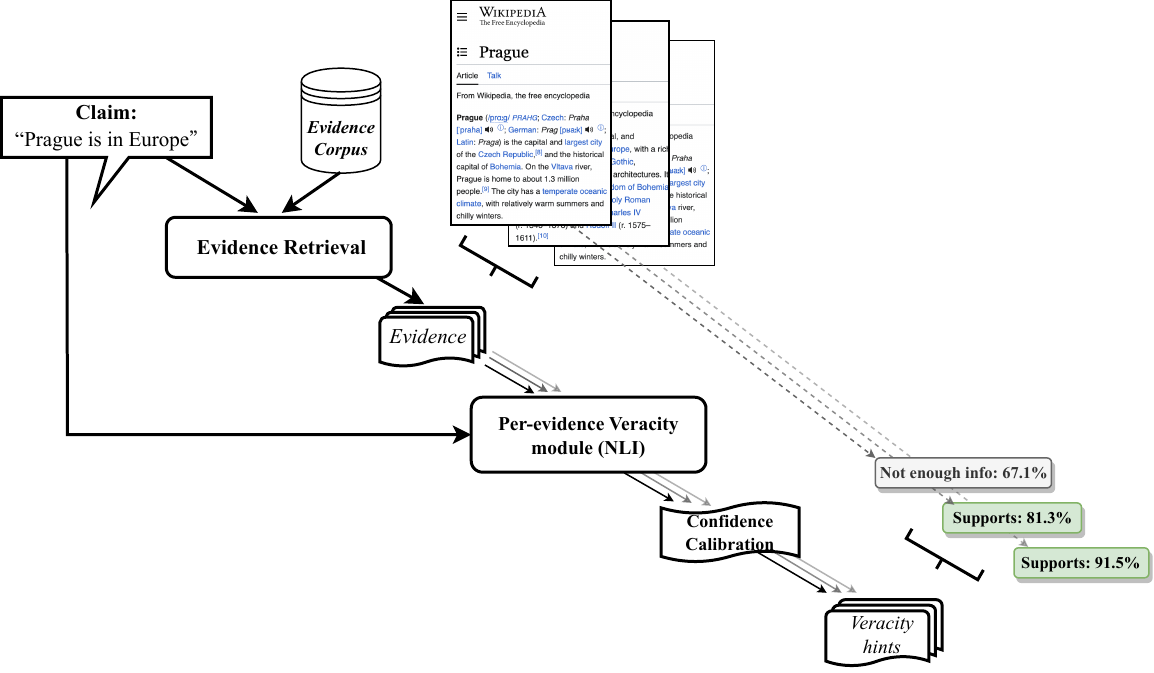}
    \caption{Automated fact-checking pipeline with paragraph-granularity retrieval and per-evidence veracity inference.}\label{fig:pipeline}
\end{figure}

Our pipeline is depicted in Figure~\ref{fig:pipeline}.
The two main modules are the \textit{evidence retrieval} (ER), which retrieves the \textit{evidence documents} subsequently classified by the \textit{veracity module}.

The task of \textit{evidence retrieval} is highly related to the \textit{document retrieval}, and while the methods are essentially the same, the difference lies in their definition of what \textit{relevant document} is.
\textit{Document retrieval} notion of relevancy is typically defined in terms of overlap in named entities, keywords, or semantic similarity.
\textit{Evidence retrieval} extends upon that, preferring documents that either help to support or refute the claim. 

The \textit{veracity module} classifies each \textit{evidence document} w.r.t. the \textit{claim} providing one of the \SUP, \REF, or \NEI labels (including class confidences).
It basically solves the Natural Language Inference (NLI) task, and we denote the data and models involved in the \textit{veracity module} as the NLI, which is common in related literature~\cite{thorne2019fever2}.
The current state-of-the-art NLI models are predominantly built using the Transformer architecture.
Unfortunately, these large-scale networks are prone to confidence miscalibration~\cite{guo2017calibration}, which we often observed in our preliminary experiments.
The miscalibration leads to non-realistic and extremely over-confident classifications (often exceeding 99\%).
To mitigate the problem, we suggest methods for calibrating the model output probabilities, and we integrate this step into our pipeline (see Section~\ref{sec:approach:nli}).

\subsection{Dataset generation}
\label{sec:approach:data}

The greatest challenge of implementing fact-checking pipelines in less-resourced languages is the unavailability of multilingual data.
Our notion of a fact-checking dataset matches that represented by \FEVER.
It consists of two main parts: the ground-truth evidence corpus composed of textual \textit{evidence documents} and a set of claims accompanied by their veracity labels and necessary evidence from this corpus.\footnote{\FEVER and some other datasets may give multiple sets of evidence per claim, providing independent annotations from multiple annotators.}

Machine translation of the available data is the most common approach in similar cases, and it can be efficiently applied to obtain NLI training data from the general fact-checking datasets, as the amount of text to be translated is limited to each claim and its textual evidence, without the need to translate the whole evidence corpus. 
Authors of this paper have already used machine translation to obtain the \FCZNLI~\cite{ullrich2023csfever} -- a Czech version of \FEVERNLI~\cite{nie2019combining}.

Unfortunately, the situation is much less convenient regarding the data needed to train the retrieval module of the fact-checking pipeline, where a substantial part of the whole evidence corpus must be translated to allow the sampling of negative examples.
With evidence corpora as large as the whole of English Wikipedia, this makes the approach needlessly costly and motivates exploration of cheaper ways to obtain retrieval data in an arbitrary language.

\subsection{Question Answering for Claim Generation}
\label{sec:approach:claims}
Rather than large-scale dataset machine translation, our pipeline builds on Question Answering for Claim Generation (\QACG)~\cite{pan2021zero} and modifies it, making the adoption of new target languages easier.
\QACG is a method that automatically generates a full fact-checking dataset of arbitrary size.
Importantly, the claims are generated directly using the corpus texts, making it easy to obtain data for specialized domains where no human-annotated data exists.
While \QACG transfer to other languages also involves machine translation, its extent is limited, and once done for the target language, it can be used to generate a fact-checking dataset from any corpus.

\begin{figure}
    \centering
    \includegraphics[width=\textwidth]{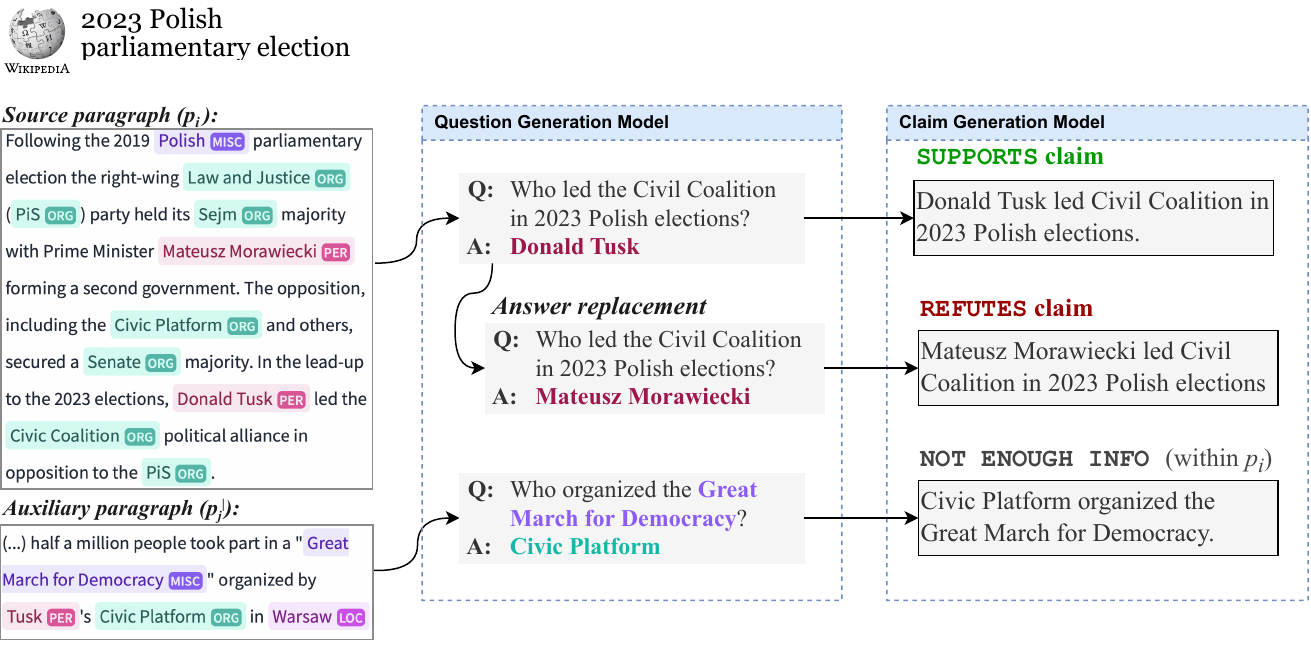}
    \caption{Our version of Question Answering for Claim Generation scheme to obtain synthetic fact-checking data. Figure, as well as approach adapted from~\cite{pan2021zero}.}\label{fig:qacg}
\end{figure}

\QACG works in three steps to generate a claim based on a source text:
\begin{enumerate}
    \item Given a context (a source document), it extracts the \textit{named entities} using a \textit{named entity recognition} (NER) method of choice.
Notably, high-quality NER solutions are already available in most examined languages. 
    \item A model trained on \textit{question answering} data generates a question for each \textit{entity}, where the \textit{entity} becomes the designated \textit{answer} to that question.
The original authors, as well as we, use the SQuAD~\cite{rajpurkar2016squad} dataset to train the \textit{question generation} (QG) model.
We use a translated version of SQuAD, where there is no existing alternative in the target language.
The QG model is trained, so it predicts a \textit{question} given the corresponding \textit{answer--context} pair.
    \item Finally, the \textit{answer--question} pair is transformed to a declarative sentence, which becomes the output \textit{claim}, using another model trained using Question to Declarative Sentence (\QATD) dataset~\cite{demszky2018transforming} (or its machine-translated version).
We call this model the \textit{claim generation} (CG) model.
\end{enumerate}
Our version of \QACG differs from~\cite{pan2021zero} in two ways: it works on the \textit{paragraph} level and simplifies the generation of \REF claims, making it language-agnostic.
The following sections cover the claim generation for all three claim types.
Figure~\ref{fig:qacg} visualizes the scheme in a simplified, digestible manner.

\subsubsection{\SUP}
First, we select $N$ random paragraphs $P=\{p_1, \ldots, p_N\}$ from the corpus.
While we work on a paragraph level, we still keep track of the \textit{original document} each paragraph belongs to, i.e., $p_i \in d(p_i)$.
We use NER to extract \textit{named entities} for each $p_i \in P$, obtaining a set of $K$ entities $A_i = \{a_1, \ldots, a_K\}$, which become answers for the following step.
The NER method provides the type of the entity denoted $\text{type}(a_j)$ as well.
In the next step, we use the QG model to obtain a question $q_{ij}$ for each input $(a_j, p_i)$ pair where $a_j \in A_i$. 
Finally, the CG model transforms the answer--question pair $(a_j, q_{ij})$ to the output claim $c_{ij}^{\text{SUP}}$. 

\subsubsection{\REF}
The \REF claims are generated based on the QG output obtained to generate the \SUP claim.
The difference lies in the CG stage where we replace the original \textit{name entity (the answer)} $a_j$ with a different entity, hence producing a likely false claim.
The original \QACG uses \textsc{sense2vec}~\cite{trask2015sense2vec} to replace the \textit{answers} with \textit{semantically aligned alternatives}.
\textsc{sense2vec} builds on \textsc{word2vec}~\cite{mikolov2013word2vec} extending it with the notion of Part-of-Speech.
Unfortunately, to our knowledge, there is neither a model like \textsc{sense2vec} nor a dataset to train such a model available for our multilingual scenario.
Therefore, we decided on a different approach to replace the answers: our method is to randomly sample a different \textit{named entity} of the same type as the original.
Formally, the CG model is fed another pair $(a_k, q_{ij})$, where $a_k \in A_i$, $a_k \neq a_j$, and $\text{type}(a_k) = \text{type}(a_j)$ getting the refuting claim $c_{ij}^{\text{REF}}$.
If such $a_k$ does not exist, the \REF claim is not generated.

\subsubsection{\NEI}
For the \NEI claims, we extract the \textit{named entities} from auxiliary paragraphs other than the \textit{source paragraph} $p_i$, but from its \textit{original document} $d(p_i)$.
Entity extraction from more than one auxiliary paragraph (two in the experiments below) aims to balance the final \QACG dataset --- we were getting too few \NEI claims given the distribution of our data.
More formally, for each evidence paragraph $p_i \in P$, we sample a set of auxiliary paragraphs from the same document $P_i' = \{p_1', \ldots, p_L'\}$, $d(p_i) = d(p_j')$ for all $p_j' \in P_i'$, where $|P_i'| = \min(|d(p_i)|, L)$. 
We extract all \textit{named entities (answers)} $A_i'$ from $P_i'$ making sure that none of them appears in $A_i$ (entities from the source paragraph $p_i$), i.e., $A_i \cap A_i' = \emptyset$.
In the next step, we use the QG model to obtain a question $q_{ij}$ for each input $(a_j', \hat{p_{ij}})$ pair, where $a_j' \in A_i'$ and $\hat{p_{ij}}$ is a concatenation of the two paragraphs: the \textit{source paragraph} $p_i$ and the auxiliary paragraph $p_j'$, from which the answer originates.
Note that we preserve the original order of paragraphs in the \textit{original document} $d(p_i)$ in the concatenation $\hat{p_{ij}}$.
Finally, using the exact same procedure as for \SUP, the CG model generates relevant claims $c_{ij}^{\text{NEI}}$ which are, however, not verifiable by $p_i$ because $a_j' \notin p_i$.

\subsection{Evidence Retrieval}
\label{sec:approach:er}
Our evidence retrieval (ER) module uses the state-of-the-art \ColBERTvT~\cite{santhanam2022colbertv2}.
The greatest motivation for choosing \ColBERTvT over the original \ColBERT~\cite{khattab2020colbert}, utilized in the preceding pipeline~\cite{ullrich2023csfever}, is not only its improved retrieval performance but also its substantially reduced memory requirements.

We train the \ColBERTvT model using the $n$-way dataset, where each claim $c$ is represented by an $(n+1)$-tuple $(c, p_1, \ldots, p_n)$.
The paragraph $p_1$ is the ground-truth \textit{evidence} (the \textit{source paragraph}) given by the dataset.
The remaining paragraphs, starting with the $p_2$, are \textit{hard negatives} selected by the \Anserini~\cite{yang2017anserini} document retrieval method ordered by the decreasing score.
Note that we use only \SUP and \REF claims for the ER model training.

\subsection{Verdict by NLI}
\label{sec:approach:nli}
Similarly to our original pipeline~\cite{ullrich2023csfever}, we use multilingual \BERT-based models as NLI classifiers for the verdict module of our pipeline.
The only change lies in the calibration, which we apply after training the model.
Our approach to mitigate the problem of overconfident predictions is the \textit{temperature scaling}~\cite{guo2017calibration}.
The advantage of the method is its simplicity, efficiency, and the fact that the calibration does not influence the actual classification but only the confidence probabilities.
This means that the \textit{confusion matrix} does not change after the model is calibrated.
The \textit{temperature scaling} algorithm optimizes a single scalar parameter $T$, which scales the \textit{logits} of the output \textit{softmax} layer of the classifier.
The optimizer minimizes the NLL loss on the validation set (denoted \dev in the following).

\section{\QACG Models and Data}\label{sec:data}
In this section, we present 1) evidence corpora datasets, 2) all \QACG training data and trained models, and finally 3) the generated \QACG datasets.

\subsection{Evidence Corpora}\label{sec:data:corpora} 

An evidence corpus is a close-world textual database of ground truth comprising all available evidence to the fact-checking pipeline.
Ideally, it should be built on top of a reliable information source.
In our past work~\cite{ullrich2023csfever}, we employed the proprietary Czech News Agency archive, where a high level of reliability is achieved by strict codex.\footnote{See Czech News Agency codex at \url{https://www.ctk.cz/o_ctk/kodex/} (Czech only).}
This paper focuses more on the technical details of building a fact-checking pipeline; hence, the specific data source and its reliability are not the highest priority.
Instead, we opted for higher reproducibility, and therefore, all evidence corpora in this study are constructed on top of public Wikipedia snapshots.
Quality-wise, Wikipedia's accuracy is mostly on par with other encyclopedic sources~\cite{casebourne, redi2019citation}.

While \FEVER dataset and its derivatives are based on significantly reduced Wikipedia snapshots, preserving only the leading section (\say{abstract}) of each article, we decided to use much larger full texts of Czech, English, Polish, and Slovak Wikipedias in most of our work.
We believe that full snapshots have significantly higher application potential, and we are interested in the scalability properties of our pipeline as well.

\subsection{Dataset generation}

We used Wikipedia snapshots from 1 August 2023 for all languages of interest.
We employed \texttt{wikiextractor}\footnote{\url{https://github.com/attardi/wikiextractor}} tool to extract plain text from each Wikipedia page.
Next, using regular expressions, we cleaned the text by removing most of the remaining HTML and JavaScript code, and we filtered out duplicate pages.
In the following step, we split the pages into paragraphs, where the successive paragraphs were merged until the total length exceeded 1000 characters.
We still call the resulting concatenated text spans the paragraphs in the following.
The merging was motivated by section titles and short paragraphs commonly present in the extracted Wikipedia pages --- in both cases, it helped to preserve context.
Additionally, the Wikipedia page title was prepended to each paragraph, aiming to provide even more contextual information.

The minimum length of a paragraph to remain in the corpus was 70 characters for all languages.
We were forced to remove 150 of the most duplicated paragraphs containing mainly general Wikipedia page status information for the Czech snapshot.\footnote{See the code supplement for more details and source code for the corpus preprocessing.} 

Table~\ref{tab:wiki} gives an overview of all four corpora denoted \CS, \EN, \PL, and \SK.
The table also provides information on \FEN~\cite{thorne2018fever} and \FCZ~\cite{ullrich2023csfever} used in some of the following experiments.
Note that both \FEVER corpora are based on older Wikipedia snapshots, taking only leading sections of each page, which roughly corresponds to our paragraph-level approach and makes comparisons reasonable.

\begin{table}[h]
    \caption{Wikipedia and \FEVER corpora statistics.}\label{tab:wiki}
    \begin{tabular}{@{}lrr@{}}
    \toprule
    language & pages & paragraphs\\
    \midrule
    \CS & \numprint{523682} & \numprint{1070289}\\
    \EN & \numprint{6320026} & \numprint{15749111}\\
    \PL & \numprint{1469845} & \numprint{2354793}\\
    \SK & \numprint{238660} & \numprint{363171}\\
    \midrule
    \FCZ & \numprint{453553} & \numprint{453553}\\
    \FEN & \numprint{5396106} & \numprint{5396106}\\
    \botrule
    \end{tabular}
\end{table}

\subsection{Named Entity Recognition}\label{sec:data:ner}

In the first stage, \QACG extracts \textit{named entities} to become the answers for the \textit{question generation} (QG) model.
We have tested multiple options for each language in preliminary experiments -- here, we summarize only the best NER models finally selected for our pipeline. 
For English and Polish, we employ default \BERT-based Stanza~\cite{qi2020stanza} toolkit models.
For Czech, we select the \SlavicBERT fine-tuned for NER task by authors of~\cite{sido2021czert}.\footnote{The \textsc{PAV-ner-CNEC} available at~\url{https://github.com/kiv-air/Czert}.}
Slovak model is the \SlovakBERT~\cite{pikuliak2022slovakbert}.\footnote{The NER fine-tuned version available at~\url{https://huggingface.co/crabz/slovakbert-ner}.}
The NER models, including model type, training data, and the number of \textit{named entity types} are summarized in Table~\ref{tab:ner}. 
\begin{table}[h]
    \caption{NER model details.}\label{tab:ner}
    \begin{tabular}{@{}lllr@{}}
    \toprule
    language & model & data & entity types\\
    \midrule
    \EN & Stanza~\cite{qi2020stanza} & OntoNotes~\cite{weischedel2012ontonotes} & 18\\
    \CS & \SlavicBERT\cite{sido2021czert} & CNEC~\cite{sevcikova2007named} & 8\\
    \PL & Stanza~\cite{qi2020stanza} & NKJP~\cite{lewandowska2012narodowy} & 6\\
    \SK & \SlovakBERT~\cite{pikuliak2022slovakbert} & WikiAnn~\cite{pan2017cross} &  4\\
    \botrule
    \end{tabular}
\end{table}
Note that the number of \textit{entity types} varies from 4 (\SK) to 18~(\EN).
The only \QACG component that works with \textit{entity types} is the \textit{claim generation} of \REF claims.
Fortunately, it does not deal with a specific value of the type but merely compares the equivalence.

In some cases, the Slovak NER model failed to detect entity span ends.
This happened for multiple successive new line characters appearing in the \textit{source text}.
Once again, we used regular expressions to detect and fix the problem.  

\subsection{Question Generation}\label{sec:data:qg}
The second stage of QACG takes each previously generated \textit{named entity}, the \textit{source paragraph} it was selected from, and generates a \textit{question} for which the  \textit{named entity} is the correct \textit{answer}.  

As in~\cite{pan2021zero}, our question generation models are trained on the question-answering datasets based on SQuAD~\cite{rajpurkar2016squad}.
SQuAD-cs v1.1 is our machine translated version of the original SQuAD v1.1, obtained using the high-quality DeepL service.\footnote{\url{http://deepl.com/}, April 28, 2023.}
The Slovak dataset SK-QuAD~\cite{hladek2023slovak} was created from scratch by human annotators.
For Polish, we have used the public dataset created using Google Translate service.\footnote{Available at \url{https://github.com/brodzik/SQuAD-PL}.} Its authors employed back translation filtering; hence, the dataset is significantly reduced in size.
We compare the datasets in Table~\ref{tab:squad}.
Note that we filtered out any unanswerable questions if present in the data.


\begin{table}[h]
    \caption{List of question answering datasets used to train the \textit{question generation} models.}\label{tab:squad}
    \begin{tabular}{@{}lllrrr@{}}
    \toprule
    language & dataset & origin & \train & \dev & \total\\
    \midrule
    \EN & SQuAD v1.1~\cite{rajpurkar2016squad} & original & \numprint{87599} & \numprint{17598} & \numprint{105197}\\
    \CS & SQuAD-cs v1.1 (ours) & DeepL & \numprint{64164} & \numprint{11722} & \numprint{75886} \\
    \PL & SQuAD-pl~\cite{brodzik2022} & Google Translate  & \numprint{34741} & \numprint{3805} & \numprint{38546} \\
    \SK & SK-QuAD~\cite{hladek2023slovak} & original & \numprint{65777} & \numprint{7808} & \numprint{73585} \\
    \botrule
    \end{tabular}
\end{table}

To train the question generation models, we have performed initial experiments (data not shown here) with three multilingual models \MBART-large-cc25~\cite{liu2020mbart}, \MTFIVE-large~\cite{xue2021mt5}, and \UMTFIVE-base~\cite{chung2022unimax}.
The differences were small, but \MTFIVE performed the best; hence, it became our choice.

We fine-tuned the \MTFIVE-large model using cross-entropy loss, AdamW~\cite{loshchilov2018decoupled} ($\beta_1 = 0.9$, $\beta_2=0.999$, and $\epsilon=10^{-8}$), with batch size 4, learning rate $3\times 10^{-5}$, using polynomial scheduler with 500 warmup steps, maximum gradient norm $0.1$, weight decay $0.1$, and label smoothing factor set to $0.1$.
The \textit{answer} and \textit{context} are separated by the \verb|</s>| separation token in the input sequence.
We trained the models for at least 85k steps, selecting the best model according to the loss computed on the \dev split (i.e., we used early stopping).
Moreover, we have merged all four language datasets to an aggregate dataset designated \SUM and trained a question generation model of the same name for 300k steps.

See the results in Table~\ref{tab:qg_models} comparing our models with the \textit{original} model presented by the authors of the original QACG study~\cite{pan2021zero} who finetuned the English monolingual \BART~\cite{lewis2020bart}.
We provide common unigram, bigram, and longest common subsequence ROUGE scores~\cite{lin2004rouge} based on the n-gram intersection of predictions with targets, as well as the model-based semantic similarity BERTScore~\cite{zhang2019bertscore}. 
None of the SQuAD datasets feature a public \test split.
Therefore, we used the \dev splits for the final evaluation.
Note that this is not a perfectly correct approach to estimate the quality of a model as the \dev splits were used in the early-stopping scheme simultaneously.
However, a thorough benchmark of different model architectures is not the main aim of this article.
Thus, we decided to keep the sizes of \train and \dev intact and use this simplified methodology. 

The results show that the \textit{original} English model is on par with our English models.
Also, we can see that the individual language models perform slightly better than the \SUM model.
Taking the almost negligible difference between models, we have opted for the \SUM in the following experiments as it is significantly more convenient to work with a single model in place of four.

\begin{table}[h]
    \caption{Comparison of \textit{question generation} models for all languages. The best results per language are highlighted in bold.}\label{tab:qg_models}
    \begin{tabular}{@{}llrrrr@{}}
    \toprule
    model & \dev & \texttt{ROUGE-1} & \texttt{ROUGE-2} & \texttt{ROUGE-L} & \texttt{BERTScore-F1}\\
    \midrule
    \EN (original)~\cite{pan2021zero} & \EN & 49.9 &    \textbf{28.4} &    46.3 &           83.7 \\
    \EN (ours)     & \EN & \textbf{50.0} &    28.1 &    \textbf{46.7} &           \textbf{84.0} \\
    \SUM           & \EN & 49.7 &    27.8 &    46.4 &           83.9 \\
    \midrule
    \CS            & \CS & \textbf{37.3} &    \textbf{19.4} &    \textbf{34.7} &           \textbf{80.0} \\
    \SUM           & \CS & 37.1 &    19.2 &    34.6 &           80.0 \\
    \midrule
    \PL            & \PL & \textbf{37.5} &    \textbf{21.4} &    \textbf{35.9} &           \textbf{81.3} \\
    \SUM           & \PL & 37.2 &    20.9 &    35.5 &           81.1 \\
    \midrule
    \SK            & \SK & \textbf{61.7} &    \textbf{45.2} &    \textbf{59.5} &           \textbf{86.8} \\
    \SUM           & \SK & 61.2 &    44.7 &    59.0 &           86.6 \\
    \botrule
    \end{tabular}
\end{table}

\subsection{Claim Generation}\label{sec:data:claim}

The last stage of the \QACG converts the \textit{answer--question} pairs to \textit{declarative sentences}, i.e., the \textit{claims}.
The authors of original \QACG approach~\cite{pan2021zero} trained the \textit{claim generation} (CG) model using the English \QATD dataset~\cite{demszky2018transforming}, which, as far as we know, has no counterpart in other languages of our interest.
Hence, we decided on the machine translation of \QATD to Czech, Polish, and Slovak using DeepL.
All translations are publicly available.
Table~\ref{tab:qatd} summarizes all \QATD datasets.

\begin{table}[h]
    \caption{\textit{Claim generation} (\QATD) datasets overview.}\label{tab:qatd}
    \begin{tabular}{@{}lllrrr@{}}
    \toprule
    language & dataset & origin & \train & \dev & \total\\
    \midrule
    \EN & \QATD~\cite{demszky2018transforming} & original & \numprint{60710} & \numprint{10344} & \numprint{71054}\\
    \CS & \QATD-cs (our) & DeepL & \numprint{60710} & \numprint{10344} & \numprint{71054} \\
    \PL & \QATD-pl (our) & DeepL & \numprint{60710} & \numprint{10344} & \numprint{71054} \\
    \SK & \QATD-sk (our) & DeepL & \numprint{60710} & \numprint{10344} & \numprint{71054} \\
    \botrule
    \end{tabular}
\end{table}

Similarly to the \textit{question generation} models, we have opted for the \MTFIVE-large~\cite{xue2021mt5} after preliminary experimentation.
We have once again merged all individual language datasets, creating the \SUM dataset.
We also compare our English models to the \textit{original} fine-tuned \BART~\cite{pan2021zero}.
The fine-tuning setup was the same as for the \textit{question generation} models (see Section~\ref{sec:data:qg}) with the only difference of training the \SUM model for a shorter period of 195k steps when it was unlikely to improve according to loss on the    \dev split. 

The results of the training are summarized in Table~\ref{tab:qatd_models}.
One can see that, once again, the differences between individual language models and the \SUM model are very small, favoring the individual language models.
The difference between our \EN model and the \textit{original} one is more pronounced in our favor.
Reasoning similarly to the \textit{question generation} models, we chose the \SUM \textit{claim generation} model for the following experiments.

\begin{table}[h]
    \caption{Comparison of \textit{claim generation} models trained on \QATD data for all languages.}\label{tab:qatd_models}
    \begin{tabular}{@{}llrrrr@{}}
        \toprule
        model & \dev & \texttt{ROUGE-1} & \texttt{ROUGE-2} & \texttt{ROUGE-L} & \texttt{BERTScore-F1}\\
        \midrule
        \EN (original)~\cite{pan2021zero} & \EN & 91.9 &    83.0 &    85.8 &           95.4 \\
        \EN (ours)     & \EN & \textbf{93.5} &    \textbf{85.9} &    \textbf{88.8} &           96.5 \\
        \SUM           & \EN & 93.5 &    85.8 &    88.7 &           \textbf{96.5} \\
        \midrule
        \CS            & \CS & \textbf{78.5} &    \textbf{63.5} &    \textbf{71.3} &           \textbf{92.2} \\
        \SUM           & \CS & 78.3 &    63.1 &    71.0 &           92.1 \\
        \midrule
        \PL            & \PL & \textbf{76.7} &    \textbf{61.7} &    \textbf{71.2} &           \textbf{92.3} \\
        \SUM           & \PL & 76.3 &    61.0 &    70.7 &           92.1 \\
        \midrule
        \SK            & \SK & \textbf{78.8} &    \textbf{64.0} &    \textbf{72.0} &           \textbf{92.2} \\
        \SUM           & \SK & 78.6 &    63.7 &    71.7 &           92.1 \\
        \botrule
        \end{tabular}
\end{table}

\subsection{\QACG Data}\label{sec:data:qacg}
In this section, we describe all our \QACG generated datasets. 
We started generating \QACG English Wikipedia data by extracting \textit{entities} from \numprint{12000} randomly sampled corpus paragraphs where, \numprint{10000} were aimed for \train split, \numprint{1000} for \dev split, and \numprint{1000} for \test split.
Each generated \textit{named entity} was successively used to generate a \textit{question}, which was converted to a \textit{claim} as described above.

We used beam search with beam size 10 for inference by both question generation and \QATD models.
As the statistics of Wikipedia texts (number and length of page paragraphs as well as the number of NERs) differ for different languages, the number of sampled paragraphs was scaled according to preliminary experiments, so the resulting number of generated claims turned out to be approximately equal.
See Table~\ref{tab:qacg_raw} for sizes of the generated raw datasets as well as their label distribution.
Note the smaller ratios for \REF and \NEI caused by the respective sampling approaches described in Section~\ref{sec:approach:claims}.

Finally, we have subsampled the generated datasets using stratified sampling to get four balanced datasets of the same size: \QACGEN, \QACGCS, \QACGPL, and \QACGSK.
The number of claims was almost halved; nevertheless, the resulting dataset sizes turned out to be appropriate in the following experiments.
Additionally, \QACG data generation is relatively cheap, so it is always possible to generate more.

In order to explore multilingual models for the fact-checking pipeline, we created two additional datasets: the \QACGMIX was uniformly sampled from all individual language datasets, preserving the overall size and label distribution, while \QACGSUM is a simple concatenation of all individual languages having a size approximately four times larger than any individual language dataset. 
See Table~\ref{tab:qacg} for the statistics of the final \QACG as well as \FEVER datasets we use in the following experiments.

Table~\ref{tab:qacg_examples_en} shows examples of \QACGCS claims generated for two evidence paragraphs.
Here, we present an English translation; the Czech original is in Table~\ref{tab:qacg_examples_cs}.
The claims generated based on the first \textit{evidence paragraph} were almost correct, except the original Czech version of the second \SUP claim, where there was an error in a preposition (see Table~\ref{tab:qacg_examples_cs}).

More interestingly, the second evidence's claims show a typical error that often happens when more than a few named entities and relations between them are present (\say{Appice \& Appice} is a malformed group name).
Moreover, the second \REF claim was generated based on the corresponding \SUP claim, replacing the name \say{Bogert} with \say{Appice}; nevertheless, the result is still factually correct.

\begin{table}[h]
    \caption{Examples of \QACGCS source evidence and the generated claims translated to English. Red color marks errors, blue the entity alternatives in \REF claim generation, and orange the error in the original Czech version not present in the translation.}\label{tab:qacg_examples_en}
    \begin{tabular}{@{}p{\textwidth}@{}}
        \toprule
\textbf{USS Indianapolis (CA-35)} \textit{(paragraph 2)}\\
\say{Indianapolis} was the last American major warship that was sunk during the war, and also the ship in which the highest number of crew members perished when it was sunk.
\textit{The wreckage.} 
On August 19, 2017, the wreckage of the ship was discovered in the Philippine Sea at a depth of 5,500 meters.
The wreckage was found by Paul Allen, who established the \say{USS Indianapolis Project} for its search. 
USS Indianapolis (CA-35) National Memorial.
The National Memorial of USS Indianapolis was officially opened on August 2, 1995. 
It is located on Canal Walk in Indianapolis. 
The limestone and granite monument depicts the heavy cruiser, with the names of the crew members inscribed below.
\vspace{1mm}

\textbf{\SUP}

The wreckage of the USS Indianapolis was discovered on August 19, 2017.

The wreckage of the USS Indianapolis was discovered \orange{in} the Philippine Sea.

\vspace{1mm}

\textbf{\REF}

The wreckage of the USS Indianapolis was discovered on \blue{August 2, 1995}.

The wreckage of the USS Indianapolis was discovered in \blue{Indianapolis}.

\vspace{1mm}

\textbf{\NEI}

The atomic bomb Little Boy was dropped on Hiroshima.

The United States Congress absolved McVay of responsibility for the loss of the ship.
\\

\midrule
\textbf{Cactus} \textit{(paragraph 2)}\\
\textit{Beck, Bogert \& Appice.}
After the breakup of the band Cactus in 1972, Bogert and Appice joined forces with Beck to form the group Beck, Bogert \& Appice.
Following one studio album ("Beck, Bogert \& Appice") and one live album ("Live In Japan," released only in Japan), the band disbanded. Their second album remains unreleased to this day, as well as the recording of the band's final concert at the Rainbow Theatre in London on January 26, 1974.
\vspace{1mm}

\textbf{\SUP}

The band Cactus disbanded in 1972.

Bogert joined forces with Beck to create the group Beck, Bogert \& Appice.

\vspace{1mm}

\textbf{\REF}

The band \red{Appice \& Appice} disbanded in 1972.

\blue{Appice} \red{joined forces with Beck to create the group Beck, Bogert \& Appice}.

\vspace{1mm}

\textbf{\NEI}

Atomic Rooster had a former member, Peter French.

Duane Hitchings played the keyboard in Cactus.
\\











\bottomrule
    \end{tabular}
\end{table}
\section{Fact-checking Pipeline Evaluation}\label{sec:exp}

In this section, we evaluate our fact-checking pipeline, comparing versions based on the  \QACG datasets generated in previous chapters as well as on the \FEVER datasets.
We evaluate both Natural Language Inference (NLI) and \textit{evidence retrieval}, employing a comprehensive assessment that includes standard metrics and human annotations.
All experiments presented in this section were performed on an AMD EPYC 7543 machine using a single 40GB NVIDIA Tesla A100.

We start by focusing on the NLI task, which aims to give the final verdict of a claim's veracity.
As discussed in Section~\ref{sec:pipeline_overview}, providing the verdict without explanation in a black-box manner is insufficient for real-world fact-checkers.
Our motivation to deal with the NLI is twofold: first, the final verdict (as well as its confidence) is likely to improve navigation in the generated fact-check reports; second, we explore how the NLI models can improve on evidence document ranking (see our previous work~\cite{ullrich2023csfever}).
Our analysis of the NLI datasets does not end with accuracy-based metrics of the trained NLI models.
We evaluate the \textit{difficulty} of the data using the novel Pointwise $\mathcal{V}$-information (PVI) approach~\cite{ethayarajh2022pvi}, and we also provide insight into claim quality and mislabel rate based on human annotations. 

The second part of this section deals with \textit{evidence retrieval} where we compare the classical BM-25 model with the state-of-the-art \ColBERTvT~\cite{santhanam2022colbertv2}.
Similarly to the NLI analysis, we contrast common \textit{document retrieval} evaluation metrics with human annotations of the retrieved evidence documents.
Moreover, we compare two approaches aimed at improving the evidence ranking: 1) the NLI reranking and 2) the keyword filtering.

Note that we only provide the complete analysis for Czech and English due to the unavailability of \FEVER data in Polish and Slovak and our limited proficiency in the latter two languages.

\subsection{Natural Language Inference}\label{sec:exp:nli}

We compare the performance of both monolingual (\QACGCS, \QACGEN, \QACGPL, \QACGSK, \FCZNLI, and \FENNLI) and multilingual (\QACGMIX and \QACGSUM) classifiers trained on all respective datasets.
The methodology to train the \FCZNLI and \FENNLI models is the same as in~\cite{ullrich2023csfever}.

All our models are based on further fine-tuning the multilingual \XLMR~\textit{large} model already fine-tuned for the SQuAD2 task.~\footnote{See \url{https://huggingface.co/deepset/xlm-roberta-large-squad2}.}
This model gave us the best results in preliminary experiments; it is also the same model used in our previous work~\cite{ullrich2023csfever}.
We employed Adam optimizer~\cite{kingba2015adam} using batch size 4, learning rate $10\times e^{-6}$, weight decay of 0.01, and maximum gradient norm 1.
We trained models for 200k steps, keeping only the best performers in terms of \dev accuracy for each model-dataset pair.\footnote{We used \QACGMIX \dev dataset for validation error estimation when training the \QACGSUM to speed up the process.}

The comparison of NLI models is shown in Table~\ref{tab:nli_langs}.
We provide F1-macro values evaluated on the \test splits of the respective datasets.
For \QACG, the \QACGSUM model dominates for all languages.
The \QACGMIX model, trained on quarter-sized data, keeps an edge over \QACGCS and \QACGSK individual language models, while the situation becomes the opposite for \QACGEN and \QACGPL.
This result supports the previous research, showing that multilingual models typically achieve better results, given enough capacity and enough data~\cite{conneau2020unsupervised}. 
\FCZNLI and \FENNLI models demonstrate significantly lower performance, which indicates that \QACG and \FEVER data do not correlate well from the NLI point of view.

The situation is very similar for \QACG models evaluated on \FCZNLI and \FENNLI, although the difference is less pronounced for Czech.
One can also see that by tuning the meta parameters, we were able to improve on results from~\cite{ullrich2023csfever}, which is most notable for \FCZNLI.  

{
\tabcolsep=0.11cm
\begin{table}[h]
    \caption{Comparison of monolingual and multilingual NLI models. F1-macro values are shown for multiple combinations of training and testing datasets. The citations mark the results taken from our previous work.}\label{tab:nli_langs}
    \begin{tabular}{@{}llr|llr@{}}
        \toprule
        \test &    \train &  F1 (\%) & \test &    \train &  F1 (\%)\\
        \midrule
          & \QACGCS                            & 77.8           &         & \QACGEN &   85.0\\
          & \QACGMIX                           & 81.3           &         & \QACGMIX &  84.5\\
\QACGCS   & \QACGSUM                           & \textbf{82.8}  & \QACGEN & \QACGSUM &  \textbf{85.9}\\
          & \FCZNLI                            & 45.8           &         & \FENNLI &   48.7 \\
          & \FCZNLI~\cite{ullrich2023csfever}  & 31.8           &         & \FENNLI~\cite{ullrich2023csfever} &  48.6 \\
        \midrule
          & \QACGPL  & 82.7          &         & \QACGSK   & 82.0 \\
\QACGPL   & \QACGMIX & 82.1          & \QACGSK & \QACGMIX  & 83.1 \\
          & \QACGSUM &\textbf{83.7}  &         & \QACGSUM  & \textbf{84.2} \\
        \midrule
          & \QACGCS                            & 46.8           &         & \QACGEN  & 54.1 \\
          & \QACGMIX                           & 47.9           &         & \QACGMIX & 51.6 \\
\FCZNLI   & \QACGSUM                           & 49.7           & \FENNLI & \QACGSUM & 49.7\\
          & \FCZNLI                            & \textbf{75.2}  &         & \FENNLI  & \textbf{84.7} \\
          & \FCZNLI~\cite{ullrich2023csfever}  & 73.7           &         & \FENNLI~\cite{ullrich2023csfever} & 75.9\\
        \bottomrule
    \end{tabular}
\end{table}
}

\subsubsection{Data Difficulty for NLI Models}\label{sec:exp:nli}

To get better insight into the difficulty of training models on all inspected datasets, we give an analysis based on the     
Pointwise $\mathcal{V}$-information (PVI)~\cite{ethayarajh2022pvi} present --- a novel method of evaluating the difficulty of a dataset w.r.t. a specific classifier model.
The PVI is based on previous work on $\mathcal{V}$-usable information (VUI) introduced in~\cite{xu2019vusable} extending the notion of VUI to individual dataset samples.
Alongside determining the difficulty of each sample, PVI can also help in finding wrongly labeled parts of a dataset, as these make the training of the model harder. 

The idea behind VUI and PVI is rooted in information theory where the well-known \textit{mutual information} is defined as:
\begin{equation}
    I(X; Y) = \mathbb{E}_{x,y \sim P(X, Y)}[\text{\footnotesize PMI}(x, y)]
\end{equation}
measures the mutual dependence between two random variables $X$ and $Y$ sampled from $\mathcal{X}$ and $\mathcal{Y}$.
The $\text{\footnotesize PMI}(x, y) \triangleq \log_2\frac{p(x, y)}{p(x) p(y)}$ is the \textit{pointwise mutual information} where in our case we deal with the dataset $\mathcal{T}^m = \{(x_i, y_i)\in\mathcal{X} \times \mathcal{Y}: i = 1, \ldots, m\}$ and our task is to predict $Y$ from $X$, e.g., labels from input text.

The problem with the mutual information is that it ignores computational constraints, so for example, the $I(X; Y)$ would not reflect the increased difficulty of the inference task when the input $X$ of an otherwise simple problem gets encrypted ($I(X; Y)$ remains constant in such case).

On the other hand, VUI works with a \textit{predictive family} $\mathcal{V} \subseteq \Omega = \{f: X \cup \varnothing \rightarrow P(Y)\}$, where the $\varnothing$ denotes a null input such as an empty input string for the text classification task.
Note that this definition of the \textit{predictive family} allows the model to ignore the inputs and base the prediction, e.g., on training data label distribution, only.
The VUI is then defined in correspondence with the \textit{mutual information}:
\begin{equation}
    I_{\mathcal{V}}(X\rightarrow  Y) = \mathbb{E}_{x,y \sim P(X, Y)}[\text{\footnotesize PVI}(x\rightarrow y)],
\end{equation}
having {\footnotesize PMI} replaced by the {\footnotesize PVI} defined as:
\begin{equation}
    \text{\footnotesize PVI}(x\rightarrow y) = - \log_2 g[\varnothing](y) + \log_2 g'[x](y),
\end{equation}
where $g' \in \mathcal{V}$ is an \textit{input-aware} model trained on $\mathcal{T}^m$ using cross-entropy loss, while $g \in \mathcal{V}$ is the \textit{null model} trained based on null input data $\mathcal{T}^m_\varnothing = \{(\varnothing, y_i)\in\{\varnothing\} \times \mathcal{Y}: i = 1, \ldots, m\}$.
Here, we adopt the notation from~\cite{ethayarajh2022pvi} where $g'[x](y)$ denotes the model's label $y$ prediction probability given the input $x$.

As for the interpretation of the {\footnotesize PVI}: the higher value indicates the less difficult sample w.r.t. the model, while lower values the more difficult one.
Negative {\footnotesize PVI} value which happens for $g[\varnothing](y) > g'[x](y)$, i.e., when the \textit{null model} $g$ assigns higher confidence to the correct class than the \textit{input-aware} classifier $g'$, implies, that it is better to classify simply based on majority than taking the input into account.

The PVI evaluation is summarized in Table~\ref{tab:pvi}.
We explore the PVI statistics for the NLI classifier introduced in the previous section (\XLMR-\textit{large}), where the \textit{null model} was trained using exactly the same approach as the \textit{input-aware model}, only having its input fixed to an empty string.
As already discussed, training NLI classifiers often ends up with uncalibrated models, inferring overly confident predictions (see Section~\ref{sec:approach:nli}); as the PVI technique is based on comparison of log confidences, we have found the calibration to be of critical importance.
Hence, we apply the \textit{temperature scaling}~\cite{guo2017calibration} for both \textit{null} and \textit{input-aware models} $g$ and $g'$.

We provide two metrics: the $\mathcal{V}$-usable information (VUI) estimate as well as the negative PVI rate (NPR) estimate, which we define as a proportion of the negative {\footnotesize PVI} values. The NPR gives insight into the percentage of \textit{suspicious} samples present in the data, where a significant number of these may be caused by mislabeled data~\cite{ethayarajh2022pvi}. 
All statistics were computed on the held-out \test splits of the respective datasets.
The table shows results for all three individual labels as well as for \total datasets.

\begin{table}[h]
    \caption{Comparison of PVI-based difficulty for \QACG and \FEVERNLI datasets on \XLMR-\textit{large} model. \textit{$\cal{V}$-usable information} (VUI) and \textit{negative PVI rate (NPR)} estimates are shown for the \test splits. The best values corresponding to the easiest data are marked in bold.}\label{tab:pvi}
    \begin{tabular}{l|rr|rr|rr|rr}
        \toprule
        \train &  \multicolumn{2}{c}{\SUP} &  \multicolumn{2}{c}{\REF} &  \multicolumn{2}{c}{\NEI} &  \multicolumn{2}{c}{\total} \\
        &  VUI &  NPR (\%) &  VUI &  NPR (\%) &  VUI &  NPR (\%) &  VUI &  NPR (\%) \\
        \midrule
        \QACGCS &     1.08 &     6.48 &     1.25 &    \textbf{12.22} &     0.09 &    40.49 &     0.81 &    19.73 \\
        \QACGEN &     \textbf{1.22} &     \textbf{5.59} &     1.19 &    16.05 &     0.37 &    24.60 &     0.93 &    15.41 \\
        \QACGPL &     1.15 &     6.58 &     1.08 &    15.94 &     0.53 &    23.90 &     0.92 &    15.47 \\
        \QACGSK &     1.18 &     6.18 &     1.01 &    18.22 &     0.49 &    26.37 &     0.89 &    16.92 \\
       \QACGMIX &     1.05 &     5.85 &     1.00 &    19.86 &     \textbf{0.67} &    \textbf{20.03} &     0.91 &    15.24 \\
       \QACGSUM &     0.99 &     7.62 &     \textbf{1.26} &    12.49 &     0.61 &    20.29 &     \textbf{0.95} &    \textbf{13.47} \\
       \midrule
           \FCZNLI &     \textbf{0.88}&    11.14 &     0.81 &    19.57 &     0.18 &    40.60 &     0.65 &    22.99 \\
           \FENNLI&     0.43 &     \textbf{4.52} &     \textbf{1.78} &    \textbf{10.65} &     \textbf{1.12} &    \textbf{28.75} &     \textbf{0.86} &     \textbf{9.53} \\
        \midrule
        \QACGCSFS &     1.09 &     6.43 &     0.80 &    23.25 &     0.33 &    34.26 &     0.74 &    21.32 \\
        \QACGENFS &     \textbf{1.10} &     \textbf{4.79} &     \textbf{0.98} &    \textbf{20.01} &     \textbf{0.52} &    \textbf{23.31} &     \textbf{0.87} &    \textbf{16.04} \\
        \bottomrule
    \end{tabular}
\end{table}

Looking at the overall results, one can see that, for the \XLMR, the \FEVERNLI datasets are, in general, somewhat more difficult than the \QACG data (lowest VUI) --- compare \QACGCS with \FCZNLI and \QACGEN with \FENNLI for the full dataset (\total).
\FCZNLI is clearly the hardest in terms of VUI; the ratio of its non-informative samples is also the largest one (NPR approaching 23\%).

\QACGSUM, which is a simple concatenation of all individual language \QACG datasets, is the least \XLMR-difficult dataset of all.
The second multilingual dataset, the \QACGMIX, preserving the size of all monolingual datasets, falls behind \QACGEN and \QACGPL, which correlates with the F1 values discussed in the previous section.

To see the impact of dataset size, we created \QACGCSFS and \QACGENFS datasets by randomly subsampling splits of the \QACGCS and \QACGEN, getting the exact split sizes and class distributions of \FCZNLI and \FENNLI.
The difference between \QACGENFS and \FENNLI becomes negligible, while even equal-sized data preserves the difference for the Czech datasets (compare \FCZNLI VUI estimate of 0.65 with 0.74 for \QACGCSFS).
This lower effect on Czech data is likely caused by the different localization approaches used when creating \FCZNLI and \QACGCSFS. 

PVI-based metrics allow for splicing the analyzed datasets; hence, we provide per-class VUI, and NPR estimates for each of \SUP, \REF, and \NEI.
One can immediately see that for all \QACG datasets, the \NEI samples are significantly the hardest.
Our first hypothesis was that this is caused by the method used to generate \NEI data, which samples other paragraphs of the same Wikipedia page as the original.
As the alternative paragraphs may still contain supporting information, the claim created by this approach may end up being mislabeled (\NEI instead of \SUP).
It turned out that the mislabeled \NEI samples are actually very sparse (see the following section); hence, our final hypothesis is that the complexity lies in the hardness of an inference from a highly relevant, yet not informative-enough, paragraph.
We plan to inspect the hypothesis in the future.

\QACGCS demonstrates the significantly lowest VUI estimate for the \NEI class (0.09), which is reflected in the confusion matrix of the corresponding NLI classifier, where about 38\% of \NEI samples are mislabeled (compared to 9\% \SUP and 20\% \REF).   

Similarly, looking at the \FENNLI data, the \REF turns out to be significantly less difficult, which can be explained by a known problem of \FEVER refute claims, where the claim creators have difficulties in creating non-trivial negatives when transforming originally \SUP claim into a \REF one~\cite{thorne2018fever}.
The problem is known as \textit{spurious cues}~\cite{derczynski2020maintaining}, and it is demonstrated by overusing words like \say{no}, \say{not}, \say{never}, or related word prefixes in the case of positive to negative claim conversion~\cite{ullrich2023csfever}.
The impact of the negating \textit{spurious cues} for for \FCZNLI was shown to be limited~\cite{ullrich2023csfever} and this is also reflected in the results. 

\subsubsection{Human Evaluation of the NLI}\label{sec:exp:human_nli}

In this section, we compare and analyze the quality of the \QACGCS, \QACGEN, \FCZNLI, and \FENNLI datasets by means of human assessment.
We selected 50 random samples per class of each dataset and annotated \textit{claim failures}, meaning incorrectly formed or grammatically wrong claim texts.
In case of a correct claim, we also checked for the correctness of its label, which should, among others, shine a light on the proportion of \textit{mislabeled'} samples and its influence on dataset difficulty, as discussed in the previous section.

Table~\ref{tab:nli_human_annotation} shows the results for each individual class as well for \total datasets.
We give the \textit{claim failure rate} --- the proportion of failed claim texts and the \textit{mislabel rate}, which denotes a proportion of mislabeled \textit{correct} claims.
All annotations were done by a single annotator (the first author of this paper, whose native language is Czech and English is his second language).
In contrast to~\cite{thorne2018fever}, we annotated the correct label with a provision for limited world knowledge beyond the context paragraph. This choice was inspired by the recognized capabilities of pre-trained Language Models (LMs) to leverage world knowledge~\cite{wang2020language}.

The \textit{claim failure rate} shows roughly double values for the \QACG data compared to \FEVERNLI, where \QACGCS reaches 19.3\%.
While such a high rate may seem to disqualify the generated data, we argue that grammatical errors caused a significant part of the failed \QACG claims; while the claim still retains the semantic quality, i.e., while incorrectly spelled, the claim is still understandable.
One can also see generally higher \textit{claim failure rates} for Czech counterparts of English data: the increase of \textit{failure rate} for the \FEVERNLI datasets can be explained by the fact that \FCZNLI is a machine-translated version of the \FENNLI, which inevitably introduce noise into data.

The cause of the quality drop for the \QACGCS compared to the \QACGEN can be two-fold.
First, the Czech language is morphologically richer and more flexible, which may lead to more probable errors or even malformed words in world cases.
This explanation is supported by our annotation experiment, where, indeed, a significant proportion of Czech claim failures demonstrated these problems.
Second, the pre-training data language distributions of the base models used for generating \QACG data are strongly skewed towards English~\cite{xue2021mt5}.

Looking at the \textit{mislabel rate} for \QACG, it is clear that unlike \SUP and \NEI, \REF has the significantly highest proportion of incorrectly labeled samples.
This is an expected outcome of selecting semantically aligned alternatives by randomly sampling \textit{named entities} of the same type from the source paragraph (see Section~\ref{sec:approach:claims}).
However, considering almost a third of \REF samples (based on correct claims) are labeled incorrectly, we plan to focus on this problem in the future, where we see model-based filtering of the \textit{semantically aligned alternatives} as a plausible solution.
One can also see that the proportion of mislabeled \NEI samples is the lowest, which was already discussed in the previous section.

For both \FEVERNLI datasets, the most mislabelings happen for the \NEI samples, which is given by the retrieval-based procedure to sample evidence for the \NEI datapoints~\cite{nie2019combining}.
\FCZNLI demonstrates more significant mislabel rates for \SUP and \REF, which can be attributed to the machine translation noise. 

\begin{table}[h]
    \caption{Human annotated \QACG and \FEVER datasets in the NLI context, 50 samples per class label (150 in total) for each dataset. F denotes the \textit{claim failure rate}, and M the \textit{mislabel rate}, both given in percent. Bold marks the best (lowest) value for each dataset.}\label{tab:nli_human_annotation}
\begin{tabular}{l|rr|rr|rr|rr}
    \toprule
    \train & \multicolumn{2}{c}{\SUP} &  \multicolumn{2}{c}{\REF} &  \multicolumn{2}{c}{\NEI} &  \multicolumn{2}{c}{\total} \\
    & F & M & F & M & F & M & F & M \\    \midrule
    \QACGCS &       18.0 &        9.8 &       24.0 &       34.2 &       \textbf{16.0} &     \textbf{9.5} &         19.3 &         17.4 \\
    \QACGEN &       10.0 &        4.4 &       16.0 &       23.8 &        \textbf{4.0} &        \textbf{2.1} &         10.0 &          9.6 \\
    \FCZNLI &        \textbf{4.0} &        \textbf{6.2} &        8.0 &       15.2 &       12.0 &       38.6 &          8.7 &         17.5 \\
    \FENNLI &        \textbf{0.0} &        \textbf{6.0} &        2.0 &       4.1 &       2.0 &       38.8 &          6.7 &         19.3 \\
    \bottomrule
    \end{tabular}
\end{table}

Table~\ref{tab:pvi_examples} shows samples with the lowest PVI value, i.e., the most \XLMR-difficult claims for the annotated portion of the \QACGEN \test split.
We give the \QACGEN label (the target), the human annotation, and the prediction of the NLI model trained on \QACGSUM.

The first example was the hardest \SUP target sample, and while the annotation and the model's prediction match the target,
one can see that the prediction confidence is low ($61.1\%$).
The second example demonstrates the case of the most difficult \REF target, which was mislabeled, as the claim is actually \SUP (based on the annotation).
The cause of the error lies in both \say{January 1, 1823} and \say{1823} named entities being extracted from the context and subsequently used as mutual alternatives in \REF generation.
The second hardest \REF target is an example of the \textit{claim failure}.
The last example demonstrates the most difficult \NEI sample of all having $\text{\footnotesize PVI} = -7.94$.
The contextual paragraph is from the same Wikipedia page as the one on which the claim was generated.
The claim and the paragraph have a common topic of discussing Republicans (Nixon mentioned in the paragraph); however, the main claim actor (Meredith) is completely missing in the context. 

\begin{table}[h]
    \caption{Examples of the annotated portion of the \QACGEN \test split having the lowest PVI values for all target classes. We include the annotated class and the \QACGSUM model predictions, including confidences.}\label{tab:pvi_examples}
    \begin{tabular}{@{}p{\textwidth}@{}}
        \toprule
\textbf{claim:} Verkhnii Turiv belonged to Turka Raion until July 2020.
\vspace{1mm}

\textbf{context:} \textit{\ldots Until 18 July 2020, Verkhnii
Turiv belonged to Turka Raion. The raion was abolished in July 2020 as part of the administrative reform of Ukraine, which reduced the number of raions of Lviv Oblast to seven. \ldots}
\vspace{1mm}

\textbf{target:} \SUP, \textbf{annotated:} \SUP, \textbf{prediction:} \SUP ($61.1\%$), $\text{\footnotesize PVI} = -4.28$
\\
\midrule

\textbf{claim:} The special elections held in Maryland were on January 1, 1823.
\vspace{1mm}

\textbf{context:} \textit{1823 Maryland's 5th congressional district special elections Special elections were held in Maryland's congressional district on January 1, 1823, \ldots}
\vspace{1mm}

\textbf{target:} \REF, \textbf{annotated:} \SUP, \textbf{prediction:} \SUP ($91.8\%$), $\text{\footnotesize PVI} = -5.36$
\\
\midrule

\textbf{claim:} The Confederate volunteer companies occupied the Civil War.
\vspace{1mm}

\textbf{context:} \textit{\ldots Local Confederate volunteer companies occupied Oglethorpe Barracks throughout American Civil War until Union General William Tecumseh Sherman captured the city in 1864. \ldots}
\vspace{1mm}

\textbf{target:} \REF, \textbf{annotated:} \FAIL, \textbf{prediction:} \SUP ($92.6\%$), $\text{\footnotesize PVI} = -3.08$
\\
\midrule

\textbf{claim:} Meredith was the candidate for the Republican nomination in 1972.
\vspace{1mm}

\textbf{context:} \textit{Gil Carmichael  Later life and death.  In 1973 Carmichael was appointed to the National Highway Safety Advisory Committee as a consolation for the Nixon administration's lack of support for him in the 1972 Senate race. He became its chairman before being made the federal commissioner for the National Transportation Policy Study Commission in 1976. He left the job in 1979. \ldots}
\vspace{1mm}

\textbf{target:} \NEI, \textbf{annotated:} \NEI, \textbf{prediction:} \REF ($97.9\%$), $\text{\footnotesize PVI} = -7.94$
\\
\bottomrule
    \end{tabular}
\end{table}


\subsection{Evidence Retrieval}\label{sec:exp:retrieval}

This section presents experiments measuring the quality of monolingual and multilingual evidence retrieval models trained on both \QACG and \FEVER data.

We compare the classical keyword-based document retrieval approach by \Anserini with neural language model-based \ColBERTvT.
The target corpora are both full Wikipedia snapshots (\CS, \EN, \PL and \SK as described in Section~\ref{sec:data:corpora}) and the older Wikipedia snapshots limited to the leading sections of each page for \FCZ~\cite{ullrich2023csfever} and \FEN~\cite{thorne2018fever}.
For \QACG datasets, we have filtered out all claim duplicates as well as all \NEI samples.
\FCZ and \FEN use the exact \train, \dev, and \test splits as in~\cite{ullrich2023csfever}.

The choice of \Anserini~\cite{yang2017anserini} was inspired by the criticism of choosing weak baselines~\cite{yang2019hype}.
We employed its Pyserini~\cite{lin2021pyserini} Python implementation in all our experiments.
For \FEN, we have set the \Anserini parameters to $k_1 = 0.6$ and $b = 0.5$; for all other datasets, we used $k_1 = 0.9$ and $b = 0.9$, which is following our previous work~\cite{ullrich2023csfever}.

The other tested model is \ColBERTvT~\cite{santhanam2022colbertv2}, which replaces \ColBERT used in our past work~\cite{ullrich2023csfever}, see Section~\ref{sec:approach} for further details.
Our slightly extended implementation is based on the reference one.\footnote{Our version of \ColBERTvT mainly extends the training loop and data import functionality. The original library is available at \url{https://github.com/stanford-futuredata/ColBERT}.}
We use \MBERT~\cite{devlin2019bert} as a backbone model for all experiments.

The \ColBERTvT was trained using the $n$-way tuples ($n=32$, see Section~\ref{sec:approach:er}), where the single positive and the remaining 31 negative paragraphs were retrieved by the \Anserini.
The total number of training triples for all datasets is summarized in Table~\ref{tab:colbert_triplets}.
All models were trained for a single epoch using cross-entropy loss with in-batch negatives,
the batch size was 128, learning rate $3\times 10^{-6}$, and we used cosine similarity metric~\cite{santhanam2022colbertv2}.
The best models were selected based on the lowest \dev split loss recorded for each of the 100 training batches.   

The results are listed in Table~\ref{tab:er_langs}, where we also include the results from~\cite{ullrich2023csfever} for \FCZ and \FEN.
Methods are compared using the Mean Reciprocal Rank (MRR) given $k \in \{1, 5, 10, 20\}$ retrieved documents.
The \train column denotes the training dataset of the corresponding model (\Anserini models are not trained prior to the indexing), while the \test column marks the corpus as well as the test split of the corresponding dataset.

Unlike in~\cite{ullrich2023csfever}, where we computed the statistics based on \textit{evidence sets} provided by multiple annotators, we simplified the methodology by considering the union of all evidence documents over all \textit{evidence sets} only.
This change was motivated by the fact that \QACG dataset claim evidence always comprises a single document, which could potentially give the \QACG an unfair advantage over \FEVER.
On the other hand, we have found that the difference in results is not significant (data not provided) as the \FEVER data involve only a limited number of the multi-hop claims (less than 13\%~\cite{jiang2020hover}) for which the evaluation of multi-document evidence sets makes sense.    

When looking at the results for the full Wikipedia corpora, it is clear that \Anserini is always outperformed by \ColBERTvT, which can be expected based on results of our previous work~\cite{ullrich2023csfever} where we employed \ColBERT.
Comparing the original \ColBERT to the newer \ColBERTvT, the latter shows a significant increase in the performance for \FCZ and \FEN.

The second observation is that both \QACGMIX and \QACGSUM multilingual models outperform all models trained on single-language datasets 
with \QACGSUM taking the lead as expected with the exception of \QACGPL, where the difference is insignificant and in favor of \QACGMIX model.
This result is in accordance with the NLI models (see Section~\ref{sec:exp:nli}) where the larger \QACGSUM data gave similarly better results than \QACGMIX, and hence we use the \QACGSUM for the analysis that follows.

Inspecting models trained on \FCZ and \FEN evaluated on full Wikipedias; we can see that their performance significantly drops when compared to all Czech and English \QACG models and \Anserini.
Also, one can see that the \QACGSUM \ColBERTvT model evaluated on \FEVER data shows similarly inferior results to the previous setup. However, the drop in performance is less pronounced, as the model at least outperforms \Anserini for both Czech and English. 

Finally, while the results cannot be directly compared for individual \QACG datasets due to significantly differing corpora, we can compare the results for \FCZ and \FEN, where \ColBERTvT qualitatively copies the results of the original \ColBERT model from~\cite{ullrich2023csfever} which performs better for \FCZ than for \FEN.
As in~\cite{ullrich2023csfever}, we explain this by the larger English corpus, which is more likely to include alternative evidence.
Interestingly, the behavior of the \QACGSUM model for \FCZ and \FEN is opposite and mimics the behavior of the keyword-oriented \Anserini. Note that the MRR scores achieved for the \Anserini and the \QACGSUM \ColBERTvT model are significantly lower.
Our hypothesis is that in this case, the models operate using lower-level syntactic features and hence are influenced by the higher flexibility of Czech and/or the noise introduced during the transfer of \FEN into \FCZ~\cite{ullrich2023csfever}.

\begin{table}[h]
    \caption{Evidence retrieval results. Comparing \Anserini and \ColBERTvT models trained on individual language, \MIX and \SUM datasets in MRR percent.}\label{tab:er_langs}
    {\centering\begin{tabular}{@{}lllrrrrr@{}}
        \toprule
        model &  \train & \test& MRR@1 &  MRR@2 &  MRR@5 &  MRR@10 &  MRR@20 \\
        \midrule
        \Anserini  &  -  & \multirow{5}{*}{\QACGCS} & 65.5 &   69.3 &   71.2 &    71.8 &    72.0 \\
        \ColBERTvT &  &  & 71.7 &   75.1 &   76.6 &    77.0 &    77.1 \\
        \ColBERTvT & \QACGMIX &  & 72.3 &   75.6 &   77.1 &    77.5 &    77.7 \\
        \ColBERTvT & \QACGSUM &  & \textbf{73.0} &   \textbf{76.0} &   \textbf{77.4} &    \textbf{77.8} &    \textbf{77.9} \\
        \ColBERTvT & \FCZ & &  20.4 &   23.3 &   25.0 &    25.5 &    25.7 \\
        \midrule
        \Anserini  &  -  & \multirow{5}{*}{\QACGEN} &  67.0 &   71.3 &   73.4 &    74.0 &    74.2 \\
        \ColBERTvT &  & & 74.1 &   77.6 &   78.9 &    79.2 &    79.4 \\
        \ColBERTvT & \QACGMIX &  & 75.6 &   79.0 &   80.3 &    80.6 &    80.7 \\
        \ColBERTvT & \QACGSUM &  & \textbf{76.5} &   \textbf{79.8} &   \textbf{81.0} &    \textbf{81.3} &    \textbf{81.4} \\
        \ColBERTvT & \FEN &  & 29.3 &   32.8 &   34.8 &    35.4 &    35.7 \\
        \midrule
        \Anserini  &  -  & \multirow{4}{*}{\QACGPL} &  59.6 &   63.6 &   65.6 &    66.1 &    66.3 \\
        \ColBERTvT & \QACGPL &  &  69.9 &   73.4 &   75.0 &    75.3 &    75.5 \\
        \ColBERTvT & \QACGMIX &  &  72.0 &   \textbf{75.5} &   \textbf{77.0} &    \textbf{77.3} &    \textbf{77.5} \\
        \ColBERTvT & \QACGSUM &  &  \textbf{72.2} &   75.5 &   76.9 &    77.3 &    77.4 \\
        \midrule
        \Anserini  &  -  & \multirow{4}{*}{\QACGSK} &  6.8 &   80.6 &   82.1 &    82.5 &    82.6 \\
        \ColBERTvT & \QACGSK &  &  80.2 &   83.4 &   84.6 &    84.9 &    85.0 \\
        \ColBERTvT & \QACGMIX &  & 80.5 &   83.4 &   84.6 &    84.9 &    85.0 \\
        \ColBERTvT & \QACGSUM &  & \textbf{81.0} &   \textbf{83.9} &   \textbf{85.0} &    \textbf{85.3} &    \textbf{85.4} \\
        \midrule
        \Anserini~\cite{ullrich2023csfever}  &  -   & \multirow{4}{*}{\FCZ} & 28.3 &   32.2 &   34.5 &    35.3 &    35.6 \\
        \ColBERT~\cite{ullrich2023csfever} & \FCZ &  &  67.2 &   71.9 &   74.3 &    74.8 &    74.9 \\
        \ColBERTvT & \FCZ &  &  \textbf{89.9} &   \textbf{91.5} &   \textbf{92.0} &    \textbf{92.1} &    \textbf{92.1} \\
        \ColBERTvT & \QACGSUM &  &  36.3 &   40.0 &   42.2 &    42.8 &    43.0 \\
        \midrule
        \Anserini~\cite{ullrich2023csfever}  &  -   & \multirow{4}{*}{\FEN} & 41.8 &   47.6 &   51.6 &    52.7 &    53.2 \\
        \ColBERT~\cite{ullrich2023csfever} & \FEN & & 62.8 &   70.0 &   72.8 &    73.5 &    73.7 \\
        \ColBERTvT & \FEN &  &  \textbf{73.1} &   \textbf{76.3} &   \textbf{77.3} &    \textbf{77.5} &    \textbf{77.6} \\
        \ColBERTvT & \QACGSUM &  &  46.6 &   50.9 &   53.2 &    53.9 &    54.2 \\
        \botrule
    \end{tabular}}
\end{table}
\subsubsection{Human Evaluation of Evidence Retrieval}\label{sec:exp:human}


This section deals with the human evaluation of evidence retrieval.
Moreover, we experiment with two augmented versions of \ColBERT: \ColBERT filtered by \Anserini denoted \ColBERTANS, and \ColBERT re-ranked by NLI denoted \ColBERTNLI (we use the shorter \ColBERT instead of \ColBERTvT for brevity in the following).

The idea of \ColBERTANS comes from the user feedback on the past versions of the FactSearch (see Section~\ref{sec:fsearch}), that a significant number of \ColBERT retrieved evidence ignores named entities and keywords present in the claim.
To mitigate the problem, \ColBERTANS takes the documents retrieved by \ColBERT and preserves only those present in top-$k$ best ranking documents retrieved by the keyword-oriented \Anserini ($k=15$ in the following experiments).

\ColBERTNLI is an alternative approach suggested, but not quantified, in our prior work~\cite{ullrich2023csfever}: we re-rank all \ColBERT retrieved documents by means of decreasing prediction confidence of the \SUM NLI model.
The confidence is based on the maximum logit value of the NLI classifier, taking only \SUP and \REF into consideration.
Moreover, we have found that the applicability of this method deteriorates fast with less-confident NLI predictions, and hence the \ColBERTNLI outputs only top-$k_2$ best-ranking results with a fairly low value of $k_2=3$ used in our experiments.

We set our first annotation experiment as follows:
\begin{enumerate}
    \item We evaluate evidence retrieval for \QACG \test claims on the corresponding full Wikipedia corpora.
    \item We report results for Czech and English due to the proficiency of the annotator in these two languages only.
The annotations were performed by the same person as in Section~\ref{sec:exp:human_nli}.
    \item The evidence retrieval \ColBERT model and the NLI model were trained on the \QACGSUM, both being the best performers in their respective categories.
    \item Each setup was evaluated based on 100 randomly sampled claims sourced from the \QACGCS and \QACGEN \test splits.
    \item The annotator was presented with two top-ranking evidence documents, where each document is supplemented by an indication of whether it constitutes the leading section of the Wikipedia page. 
    \item Each evidence document is evaluated either as \textit{relevant} or \textit{non-relevant}.
The annotator should mark the document as relevant if the document directly supports or refutes the claim or if it may strongly help with (dis)proval of the claim.
Consequently, the leading section corresponding to an important named entity present in the claim was almost always approved as relevant.  
\end{enumerate}

The results are summarized in Table~\ref{tab:human_qacg}.
We provide MRR and precision based on \textit{annotated} relevance labels as well as on the \textit{target} labels given by the respective \test splits so we can directly compare to the Table~\ref{tab:er_langs}.

One can immediately see that the \textit{annotated} MRR@2 and P@2 are significantly higher than the corresponding \textit{targets}.
This can be easily explained by the fact that each claim of the \QACG data is based on a single document (the \textit{target}), while there are often multiple relevant evidence documents in the corpus (even more probable for the full Wikipedia corpora).

Concerning the retrieval methods, \Anserini is always superior to the models trained on \FEVER, which was already observed in the previous section.
\ColBERTNLI is the worst performing method with few exceptions where it outperforms \ColBERT in precision.
Nevertheless, this happens mostly for the \textit{targets}, meaning \ColBERTNLI retrieved documents are generally not preferred by the annotator.
The annotator's observation was that reranking by the NLI model leads to even fewer retrieved documents containing any relevant named entities.

\ColBERTANS is prefered by \textit{annotations}, where the \train dataset matches the \test one (for the \QACGSUM model).
A single exception happens for the MRR of \QACGSUM model tested on \QACGEN, where \ColBERT slightly outperforms \ColBERTANS; the precision is, on the other hand, more significantly in favor of the \ColBERTANS.
Notably, the similar improvement achieved by \Anserini filtering is not observed for the \textit{targets}, which can be explained by the annotator's higher preference for correct named entities and keywords in the retrieved documents.

The ranking of both MRR and precision is mostly preserved between \textit{targets} and \textit{annotations} for the lower quality \FEVER models with a sole and less significant exception of the \FEN-trained model MRR, where \ColBERT and \ColBERTNLI order is swapped.
For the high quality \QACG-trained models, the method ranking between \textit{targets} and human \textit{annotations} widely differs (with the exception of the MRR for \QACGSUM model tested on \QACGEN \test split).



The second annotation experiment aims to inspect generalization capabilities.
Its setup is exactly the same as in the previous case.
The only difference lies in using \FCZ and \FEN \test split claims instead of the corresponding \QACG ones.
This means we have no \textit{targets} for comparison, so we only provide \textit{annotation}-based results.
The motivation behind this choice was that the \FEVER claims, while created in an artificial process, are constructed by humans, hence more likely akin to real-world claims.
The decision to retrieve from the full Wikipedia corpora was once again motivated by testing in the more realistic scenario as \FEVER corpora are limited to Wikipedia texts truncated to leading sections only.



\begin{table}[h]
    \caption{Evidence retrieval for \CS and \EN Wiki corpora and \QACG \test splits: human evaluation. The evidence relevancy is compared for the human \textit{annotations} and the \QACG \test split labels (\textit{targets}).} \label{tab:human_qacg}
    \begin{tabular}{@{}lll|lr|lr@{}}
    \toprule
    \test & \train & model & \multicolumn{2}{c}{MRR@2 (\%)}  &  \multicolumn{2}{c}{P@2 (\%)}  \\
    & & & annotated & target & annotated & target\\
    \midrule

    \multirow{7}{*}{\QACGCS} & - & \Anserini & 86.5 & 69.3 & 66.0 & 36.6 \\
    \cmidrule{2-7}
    & \multirow{3}{*}{\FCZ} & \ColBERT & 50.0 & 23.3 & 33.5 & 13.1 \\
    &                       & \ColBERTANS & \textbf{54.5} & \textbf{39.9} & \textbf{38.5} & \textbf{21.7} \\
    &                       & \ColBERTNLI & 50.0 & 27.8 & 34.0 & 14.4 \\
    \cmidrule{2-7}
    & \multirow{3}{*}{\QACGSUM} & \ColBERT &  86.5 & \textbf{76.0} & 63.5 & \textbf{39.5} \\
    &                           & \ColBERTANS & \textbf{89.0} & 68.7 & \textbf{66.5} & 35.2 \\
    &                           & \ColBERTNLI & 84.0 & 67.2 & 63.0 & 37.5 \\

    \midrule

    \multirow{7}{*}{\QACGEN} & \Anserini & - & 90.5 & 71.3 & 71.0 & 37.9 \\
    \cmidrule{2-7}
    & \multirow{3}{*}{\FEN} & \ColBERT & 65.0 & 32.8 & 44.5 & 18.2 \\
    &                       & \ColBERTANS & \textbf{76.5} & \textbf{52.7} & \textbf{55.5} & \textbf{28.4} \\
    &                       & \ColBERTNLI & 63.0 & 36.2 & 46.0 & 19.2 \\
    \cmidrule{2-7}
    & \multirow{3}{*}{\QACGSUM} & \ColBERT & \textbf{95.0} & \textbf{79.8} & 68.5 & \textbf{41.5} \\
    &                           & \ColBERTANS & 94.0 & 72.8 & \textbf{73.0} & 37.3 \\
    &                           & \ColBERTNLI & 81.0 & 66.5 & 65.5 & 37.9 \\

    \botrule
    \end{tabular}
\end{table}

The results are shown in Table~\ref{tab:human_fever}.
Similar to the previous experiments, one can see that English models outperform the Czech models.
Another observation is that \ColBERT-based methods win only in cases where \train dataset matches the \test one.
Nevertheless, in both such cases, the performance gain achieved is significant.
\Anserini once again outperforms the models trained on \QACG.
The only exception is \QACGSUM \train and \FEN \test for which the \ColBERTANS is on par: precision is 58.0\% for \ColBERTANS compared to 57.0\% for \Anserini, and opposite ranking for the MRR (66.5\% for \ColBERTANS, 67.5\% for \Anserini).
As in the previous experiment, \ColBERTNLI is the worst performer in all cases.
\ColBERTANS brings a reasonable edge over \ColBERT for \FEN \train and \test splits; however, the \Anserini filtering harms the performance for the \FCZ in terms of MRR (precision is preserved for both).

In summary, one can see that, following the previous analysis, \FEVER and \QACG data distributions differ significantly, and hence, the resulting models provide only limited generalization abilities.
Comparing the \textit{annotations} to the \textit{targets} reveals the importance of retrieving documents based on correct named entities and/or keywords, making the \Anserini baseline even stronger in practice.

\begin{table}[h]
    \caption{Evidence retrieval for \CS and \EN Wiki corpora and \FEVER \test splits: human evaluation.}\label{tab:human_fever}
    \begin{tabular}{@{}lll|l|l@{}}
    \toprule
    \test & \train & model & MRR@2 (\%) & P@2 (\%) \\
    \midrule
    \multirow{7}{*}{\FCZ} & - & \Anserini &  61.0 & 46.5 \\
    \cmidrule{2-5}
    & \multirow{3}{*}{\FCZ} & \ColBERT & \textbf{78.5} & \textbf{49.0} \\
    &                                  & \ColBERTANS & 75.5 & \textbf{49.0} \\
    &                                  & \ColBERTNLI & 59.5 & 40.5 \\
    \cmidrule{2-5}
    & \multirow{3}{*}{\QACGSUM} & \ColBERT    & 52.5 & 37.0 \\
    &                           & \ColBERTANS & \textbf{54.5} & \textbf{41.5} \\
    &                           & \ColBERTNLI & 46.5 & 31.5 \\

    \midrule
    
    \multirow{7}{*}{\FEN} & - & \Anserini & 67.5 & 57.0\\
    \cmidrule{2-5}
    & \multirow{3}{*}{\FEN} & \ColBERT & 72.0 & 56.5 \\
    &                       & \ColBERTANS & \textbf{74.5} & \textbf{60.0} \\
    &                       & \ColBERTNLI & 65.0 & 52.5 \\
    \cmidrule{2-5}
    & \multirow{3}{*}{\QACGSUM} & \ColBERT & 60.0 & 49.5\\
    &                           & \ColBERTANS & \textbf{66.5} & \textbf{58.0} \\
    &                           & \ColBERTNLI & 57.0 & 46.0 \\

    \botrule
    \end{tabular}
\end{table}

\section{FactSearch Application}\label{sec:fsearch}

The previous section dealt with both quantitative and qualitative evaluation of our pipeline.
This section describes the next step, where we present a prototype of the FactSearch web application aimed at getting preliminary user feedback.

\begin{figure}
    \centering
    \includegraphics[width=\textwidth]{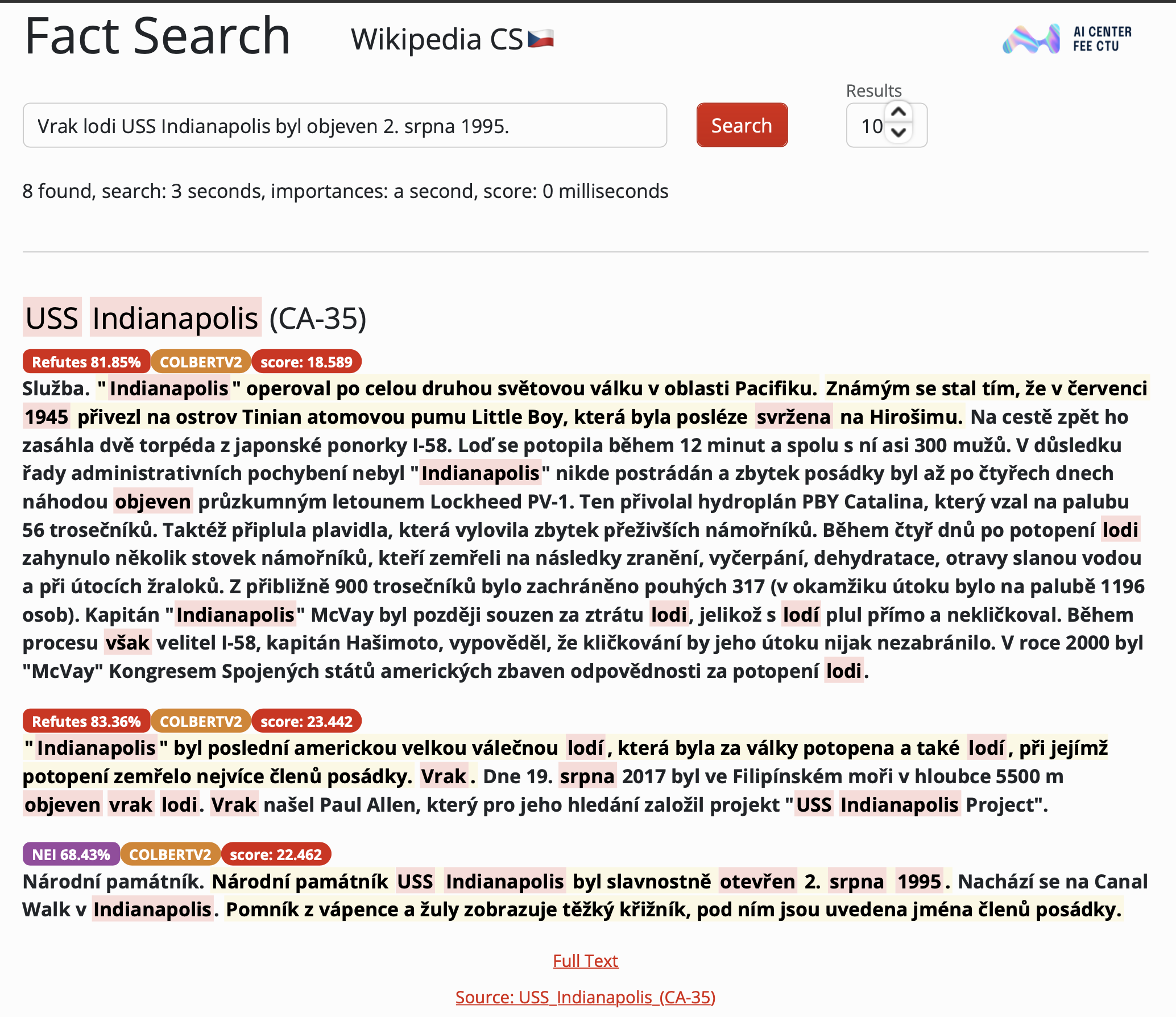}
    \caption{FactSearch application implementing the pipeline for Czech Wikipedia.}\label{fig:factsearch}
\end{figure}

Figure~\ref{fig:factsearch} shows the FactSearch interface, where a user enters a claim into the entry field and is given results along with several statistics.
The veracity of the retrieved evidence paragraphs is analyzed using the NLI model, and the most likely class and classification confidence is provided.
We also give information on the evidence retrieval method used (\ColBERTvT in this case) and the score of the evidence retrieval.
The evidence paragraphs are sorted by decreasing retrieval score, except for paragraphs belonging to the same source document (Wikipedia page), as shown in the example.
Users can display the full text of the \textit{source document} either using the \textit{"Full Text"} link or via a link to the actual Wikipedia page.

Additionally, we emphasize words similar to those in the claim (pink background) to improve orientation in the retrieved texts.
The procedure is based on matching claim and paragraph words using Jaro--Winkler similarity~\cite{winkler90string}.
We emphasize only words longer than three characters having similarity above the threshold of 0.8.
This seemingly very simple approach works surprisingly well for all four languages.

We have received initial feedback by six journalists based in AFP\footnote{\url{https://afp.com}}, Demagog.cz\footnote{\url{https://demagog.cz}}, and Czech Radio\footnote{\url{https://portal.rozhlas.cz}}.

The users reported the NLI module to be the most problematic of all, as the veracity classification was noisy: the classification of two similar texts was often classified as contradictory.
We hypothesize that it is mainly caused by the noise in the NLI portion of the data.
Also, the experiments in the previous section showed limited transferability of models trained on \FEVER and \QACG data.

The evidence retrieval performance of FactSearch was reported to be problematic only by some of the users.
Our inspection, however, showed that this problem was caused mainly by overfitting our ER models to the exact claim input form.
Notably, all the \QACG-generated training claims had a form of a declarative sentence, commencing with a capital letter and terminating with a period.
We found that using other forms, such as all lower-case letters and no punctuation, led to a significant drop in the retrieval quality.
Fortunately, addressing the issue of enhancing ER data for input invariance is a manageable task, and we intend to tackle this challenge in the next iteration of the pipeline. 

The user interface and the highlighting of the words matched with the claim were generally well-received.

Overall, the idea of a tool based on the principles of FactSearch was accepted very positively if the tool was deployed on top of more interesting text corpora, where the reviewers suggested the AFP archive.

\section{Conclusion}\label{sec:conclusion}
In this article, we introduced an automated fact-checking pipeline with a primary emphasis on facilitating implementation in languages with fewer available resources.
Our strategy involves synthetically generated data to circumvent the complexities and expenses associated with annotating data required for training both the \textit{evidence retrieval} and \textit{claim veracity evaluation} modules of the pipeline working on the level of paragraphs.

Our \textit{evidence retrieval} module is based on the \ColBERTvT late-interaction document retrieval method~\cite{santhanam2022colbertv2}, while the \textit{claim veracity module} is implemented using the NLI classifier, namely \XLMR~\cite{conneau2020unsupervised}.

We have modified the \QACG~\cite{pan2021zero} data generating method so that only \textit{question generation}, \textit{claim generation} training data, and a \textit{named entity recognition} model are needed to deploy the pipeline.
Both datasets have moderate size, facilitating the machine translation to the \textit{target language}.

We provide all \QACG data and models needed to generate data in Czech, English, Polish, and Slovak.
Additionally, we release extensive fact-checking training datasets derived from Wikipedia for all four languages.

We have extensively tested both pipeline modules, including a comparison with their Czech and English \FEVER-based alternatives.
For the NLI classifier, the best results were achieved by the multilingual model trained on the sum of the generated \QACG data for all four languages (\QACGSUM).
We have found that the transfer between \QACG and \FEVER domains is limited, i.e., the prediction of an NLI model trained in one does not perform well for the claims of the other.
To shed some light on dataset differences, we performed further analyses within the framework of Pointwise $\mathcal{V}$-information (PVI) as well as human annotations.

The PVI enables the assessment of dataset difficulty on both a per-class and per-sample basis in relation to a particular classifier.
We have illustrated variations in per-class difficulties and linked them to distinct approaches in claim generation within \QACG and \FEVER datasets.
We have also presented examples of the most difficult \QACG claims, i.e., the claims having the lowest PVI values in each class.
These samples demonstrate problems happening during the \QACG data generation process, namely the \textit{claim failure} (wrongly formed claims) and \textit{mislabeled claims} caused by errors in \textit{semantically aligned replacement}.

To get further insight into the data, we have annotated a sample of the Czech and English \QACG and \FEVERNLI datasets.
The results show higher \textit{failure rate} for \QACG data compared to \FEVER.
Nevertheless, a significant portion of unsuccessful claims resulted from minor grammatical errors that had a minimal impact on the meaning of the claims (we intend to quantify this precisely in future research).
The overall \textit{mislabel rate} is comparable for \QACGCS and \FCZNLI; however, it is about half for the English \QACGEN compared to \FENNLI.

Where the original authors of \QACG focused solely on NLI~\cite{pan2021zero}, we continued analyzing the properties of \QACG data w.r.t. \textit{evidence retrieval}.
The results once again show that the multilingual models trained using the aggregated \QACG data of all four languages work the best.
Similarly to the NLI, we experienced limited transferability between \FEVER to \QACG domains.
Interestingly, while \FEVER-trained models significantly failed for \QACG claims, the opposite direction was more favorable to our version of \QACG: the \QACG-trained \ColBERTvT models were at least able to outperform the \Anserini BM-25 baseline.
Overall, \ColBERTvT is an improvement over its predecessor, which was part of our past pipeline~\cite{ullrich2023csfever}.

Similar to the NLI, we relied on human annotations to evaluate the relevance of retrieved evidence, considering all four combinations of \QACG and \FEVER for training and testing.
We also included two modifications of the \ColBERTvT retrieval procedure: reranking by NLI (\ColBERTNLI) and \Anserini filtering (\ColBERTANS).
The former was proposed in our prior work~\cite{ullrich2023csfever}. 
The latter was inspired by user feedback on the initial version of our pipeline, where the retrieved evidence occasionally overlooked crucial named entities or keywords.
Although \ColBERTNLI generally showed subpar results, \ColBERTANS demonstrated improved performance compared to plain \ColBERT in the majority of cases.
This enhancement was particularly noticeable in human evaluations, where there was a preference for the inclusion of essential named entities.

Finally, we developed the FactSearch application utilizing the pipeline.
The initial feedback suggests the concept's utility while highlighting concerns about the practical accuracy of the \textit{claim veracity evaluation}, i.e., the NLI model.

\subsection{Future Work}
\begin{itemize}
    \item Analyzing our \QACG generated data, we found that their most limiting property is the high misclassification rate of the \REF samples, which introduces noise in training the NLI models.
Our main concern in the future is to reduce the extent of this problem by improving the selection of \textit{semantically aligned alternatives}.
One possible solution is the syntactic and semantic deduplication of the extracted named entities. 
    \item Our main focus when designing the pipeline was on ease of implementation to new languages with no access to human-annotated data.
If such data is available, even in a limited size, we can further fine-tune our \QACG models as suggested by the authors of the original \QACG~\cite{pan2021zero}.
    \item The current method of generating data does not involve multi-hop claims, i.e., claims where more than a single evidence document is needed to assess the claim veracity.
We plan to address this deficiency by extending \QACG by another retrieval module to generate chains of related named entities.
    \item The \QACG is a two-stage process: the question generation and the successive claim generation, with both stages in need of some training data in target language. Could a simpler scheme, such as a single-sentence abstractive summarization~\cite{narayan-etal-2018-dont} be used to generate claims directly from text instead, leading to further reduction of the data requirements? A further exploration of such scheme is desirable to establish its benefits and tradeoffs.
    \item The authors of \ColBERTvT utilized the earlier version of \ColBERT~\cite{khattab2020colbert} to bootstrap the generation of the $n$-way training tuples.
In contrast, our approach in this paper was streamlined, opting for the BM-25 document retrieval method (\Anserini).
We intend to experiment with multiple stages of \ColBERTvT $n$-way tuple generation, with the goal of obtaining an even more robust representation of hard negatives in the training data.
    \item We plan to extend the deployment of the pipeline to corpora beyond Wikipedia.
Currently, we are conducting preliminary experiments with several reliable Czech news sources.
Informal discussions with journalists have suggested that an intriguing avenue could involve fact-checking claims against sources of disinformation.
\end{itemize}

\bmhead{Acknowledgments}
This article was produced with the support of the Technology Agency of the Czech Republic under the ÉTA Programme, project TL05000057. The access to the computational infrastructure of the OP VVV funded project CZ.02.1.01/0.0/0.0/16\_019/0000765 ``Research Center for Informatics'' is also gratefully acknowledged.

\bmhead{Competing Interests}
The authors declare that no conflict of interest exists.

\bmhead{Data Availability}
All data, models, and source code to reproduce our work are available at: \url{https://github.com/aic-factcheck/multilingual-fact-checking}. 

\begin{appendices}
\section{Additional Dataset Statistics}\label{sec:appendix_a}

Table~\ref{tab:qacg_raw} summarizes the sizes and label distribution of the raw generated \QACG data.
See Section~\ref{sec:data:qacg} for the details.

\begin{table}[h]
    \caption{\QACG generated raw claims and their label distribution.}\label{tab:qacg_raw}
    \begin{tabular}{lrrrr}
    \toprule
    language &  \SUP (\%) & \REF (\%) & \NEI (\%) & \total \\
    \midrule
      \CS & 38.3 & 36.5 & 25.2 & \numprint{618687} \\
      \EN & 39.0 & 34.9 & 26.1 & \numprint{628622} \\
      \PL & 39.0 & 34.9 & 26.1 & \numprint{628622} \\
      \SK & 40.2 & 38.6 & 21.2 & \numprint{556873} \\
    \bottomrule
    \end{tabular}
\end{table}

Table~\ref{tab:qacg} presents statistics of the balanced \QACG datasets and \FEVER datasets.
See Section~\ref{sec:data:qacg} for the details.

\begin{table}[h]
    \caption{\QACG balanced datasets and \FEVER datasets statistics.}\label{tab:qacg}
    \begin{tabular}{@{}lrrrrrrr@{}}
    \toprule
    dataset & \train & \dev & \test & \total & \SUP & \REF & \NEI\\
    \midrule
    \QACGCS & \numprint{295209} &  \numprint{30087} &  \numprint{28440} &  \numprint{353736} & \numprint{117912} & \numprint{117912} & \numprint{117912} \\
    \QACGEN & \numprint{295209} &  \numprint{30087} &  \numprint{28440} &  \numprint{353736} & \numprint{117912} & \numprint{117912} & \numprint{117912} \\
    \QACGPL & \numprint{295209} &  \numprint{30087} &  \numprint{28440} &  \numprint{353736} & \numprint{117912} & \numprint{117912} & \numprint{117912} \\
    \QACGSK & \numprint{295209} &  \numprint{30087} &  \numprint{28440} &  \numprint{353736} & \numprint{117912} & \numprint{117912} & \numprint{117912} \\
    \QACGMIX & \numprint{295209} &  \numprint{30087} &  \numprint{28440} &  \numprint{353736} & \numprint{118039} & \numprint{117640} & \numprint{118057} \\
    \QACGSUM & \numprint{1180836} & \numprint{120348} & \numprint{113760} & \numprint{1414944} & \numprint{471648} & \numprint{471648} & \numprint{471648} \\
    \midrule
    \FCZ & \numprint{107330} &  \numprint{9999} &  \numprint{9999} &  \numprint{127328} & \numprint{60208} & \numprint{24815} & \numprint{42305} \\
    \FEN & \numprint{145449} &  \numprint{19998} &  \numprint{19998} &  \numprint{185445} & \numprint{100033} & \numprint{49773} & \numprint{55637} \\
    \botrule
    \end{tabular}
\end{table}

Table~\ref{tab:colbert_triplets} gives an overview of the number of \ColBERTvT training tuples.
See Section~\ref{sec:exp:retrieval} for the details.
\begin{table}[h]
    \caption{\ColBERTvT: the number of \QACG and \FEVER training tuples.}\label{tab:colbert_triplets}
    \begin{tabular}{lr}
    \toprule
    dataset & tuples \\
    \midrule
      \QACGCS & \numprint{225684} \\
      \QACGEN & \numprint{223456} \\
      \QACGPL & \numprint{219620} \\
      \QACGSK & \numprint{219962} \\
      \QACGMIX & \numprint{223385} \\
      \QACGSUM & \numprint{888722} \\
    \midrule
      \FCZ & \numprint{84803} \\
      \FEN & \numprint{123139} \\
    \bottomrule
    \end{tabular}
\end{table}

Table~\ref{tab:qacg_examples_cs} presents the original Czech source evidence and \QACG-generated claims discussed in Section~\ref{sec:data:qacg}.
The English version of the same was shown in Table~\ref{tab:qacg_examples_en}.
The wrong preposition \say{na} (\say{on} in English) in the second \SUP claim example for the \textit{USS Indianapolis} evidence.
The correct Czech preposition in this case would be \say{v} (\say{in}).
This error has not propagated to the English translation in Table~\ref{tab:qacg_examples_en}.
Unlike \QACGEN, similar mistakes happen relatively commonly for all West Slavic languages due to the declension.
 
\begin{table}[h]
  \caption{Examples of \QACGCS dataset claims. Red color marks errors, and blue the entity alternatives in \REF claim generation.}\label{tab:qacg_examples_cs}
  \begin{tabular}{@{}p{\textwidth}@{}}
      \toprule
\textbf{USS Indianapolis (CA-35)} \textit{(paragraph 2)}\\
\say{Indianapolis} byl poslední americkou velkou válečnou lodí, která byla za války potopena a také lodí, při jejímž
potopení zemřelo nejvíce členů posádky.
\textit{Vrak.}
Dne 19. srpna 2017 byl ve Filipínském moři v hloubce 5500 m objeven vrak lodi.
Vrak našel Paul Allen, který pro jeho hledání založil projekt \say{USS Indianapolis Project}.
USS Indianapolis (CA-35) Národní památník.
Národní památník USS Indianapolis byl slavnostně otevřen 2. srpna 1995.
Nachází se na Canal Walk v Indianapolis.
Pomník z vápence a žuly zobrazuje těžký křižník, pod ním jsou uvedena jména členů posádky.
\vspace{1mm}

\textbf{\SUP}

Vrak lodi USS Indianapolis byl objeven 19. srpna 2017.

Vrak lodi USS Indianapolis byl objeven \red{na} Filipínském moři.

\vspace{1mm}

\textbf{\REF}

Vrak lodi USS Indianapolis byl objeven \blue{2. srpna 1995}.

Vrak lodi USS Indianapolis byl objeven v \blue{Indianapolis}.

\vspace{1mm}

\textbf{\NEI}

Atomová puma Little Boy byla svržena v Hirošimu.

Kongres Spojených států osvobodil McVayho z odpovědnosti za ztrátu lodi.\\

\midrule













\textbf{Cactus} \textit{(paragraph 2)}\\
\textit{Beck, Bogert \& Appice.}
Po rozpadu skupiny Cactus v roce 1972, Bogert a Appice se spojili s Beckem, aby vytvořili skupinu Beck, Bogert \& Appice.
Po jednom studiovém album (\say{Beck, Bogert \& Appice}) a jednom živém albu (\say{Live In Japan}, vydaném pouze v Japonsku) se skupina rozpadla. Jejich druhé album zůstalo nevydáno dodnes, stejně tak jako záznam posledního koncertu skupiny, který se konal v londýnském Rainbow Theatre 26. ledna 1974.
\vspace{1mm}

\textbf{\SUP}
Skupina Cactus se rozpadla v roce 1972.

Bogert se spojil s Beckem a vytvořil skupinu Beck, Bogert \& Appice.

\vspace{1mm}

\textbf{\REF}
Skupina \red{Appice \& Appice} se rozpadla v roce 1972.

\blue{Appice} \red{se spojil s Beckem a vytvořil skupinu Beck, Bogert \& Appice}.

\vspace{1mm}

\textbf{\NEI}

Atomic Rooster měl bývalého člena Petera Frenche.

Duane Hitchings hrál na klávesy v Cactus.
\\
\bottomrule
  \end{tabular}
\end{table}
\end{appendices}


\bibliography{main}

\end{document}